\documentclass[runningheads]{llncs}

\usepackage[mobile]{eccv}

\usepackage{eccvabbrv}

\usepackage{graphicx}
\usepackage{booktabs}

\usepackage[accsupp]{axessibility}  
\usepackage{colortbl}
\usepackage{xcolor}
\usepackage{array}
\usepackage{utfsym}
\usepackage{soul}

\usepackage{hyperref}

\usepackage{orcidlink}

\usepackage{multirow}
\usepackage{bbding}
\usepackage{roboto}
\usepackage{makecell}
\usepackage{float}

\begin{document}
\def\methodNAME{PartGLEE\xspace}
\definecolor{darkgreen}{rgb}{0,0.7,0}
\definecolor{lightgray}{RGB}{158, 153, 153}
\title{PartGLEE: A Foundation Model for Recognizing and Parsing Any Objects}

\titlerunning{PartGLEE}

\author{Junyi Li\inst{1*} \and
Junfeng Wu\inst{1*} \and
Weizhi Zhao\inst{1} \and
Song Bai\inst{2} \and 
Xiang Bai\inst{1\dag}
}

\authorrunning{J. Li, J. Wu, et al.}

\institute{
Huazhong University of Science and Technology \and ByteDance Inc.
}

\maketitle

\begin{abstract}
\def\thefootnote{*}\footnotetext{Equal Technical Contribution. Work done during Junfeng's internship at ByteDance. $^\dagger$Correspondence to Xiang Bai $<$\url{xbai@hust.edu.cn}$>$.}
We present \methodNAME, a part-level foundation model for locating and identifying both objects and parts in images. Through a unified framework, \methodNAME accomplishes detection, segmentation, and grounding of instances at any granularity in the open world scenario. Specifically, we propose a Q-Former to construct the hierarchical relationship between objects and parts, parsing every object into corresponding semantic parts.
By incorporating a large amount of object-level data, the hierarchical relationships can be extended, enabling \methodNAME to recognize a rich variety of parts.
We conduct comprehensive studies to validate the effectiveness of our method, \methodNAME achieves the state-of-the-art performance across various part-level tasks and obtain competitive results on object-level tasks. 
The proposed \methodNAME significantly enhances hierarchical modeling capabilities and part-level perception over our previous GLEE model.
Further analysis indicates that the hierarchical cognitive ability of \methodNAME is able to facilitate a detailed comprehension in images for mLLMs. The model and code will be released at \url{https://provencestar.github.io/PartGLEE-Vision/}.


\end{abstract}
 
\section{Introduction}
\label{sec:intro}

\begin{figure*}[tb]
\centering
\includegraphics[width=0.99 \linewidth]{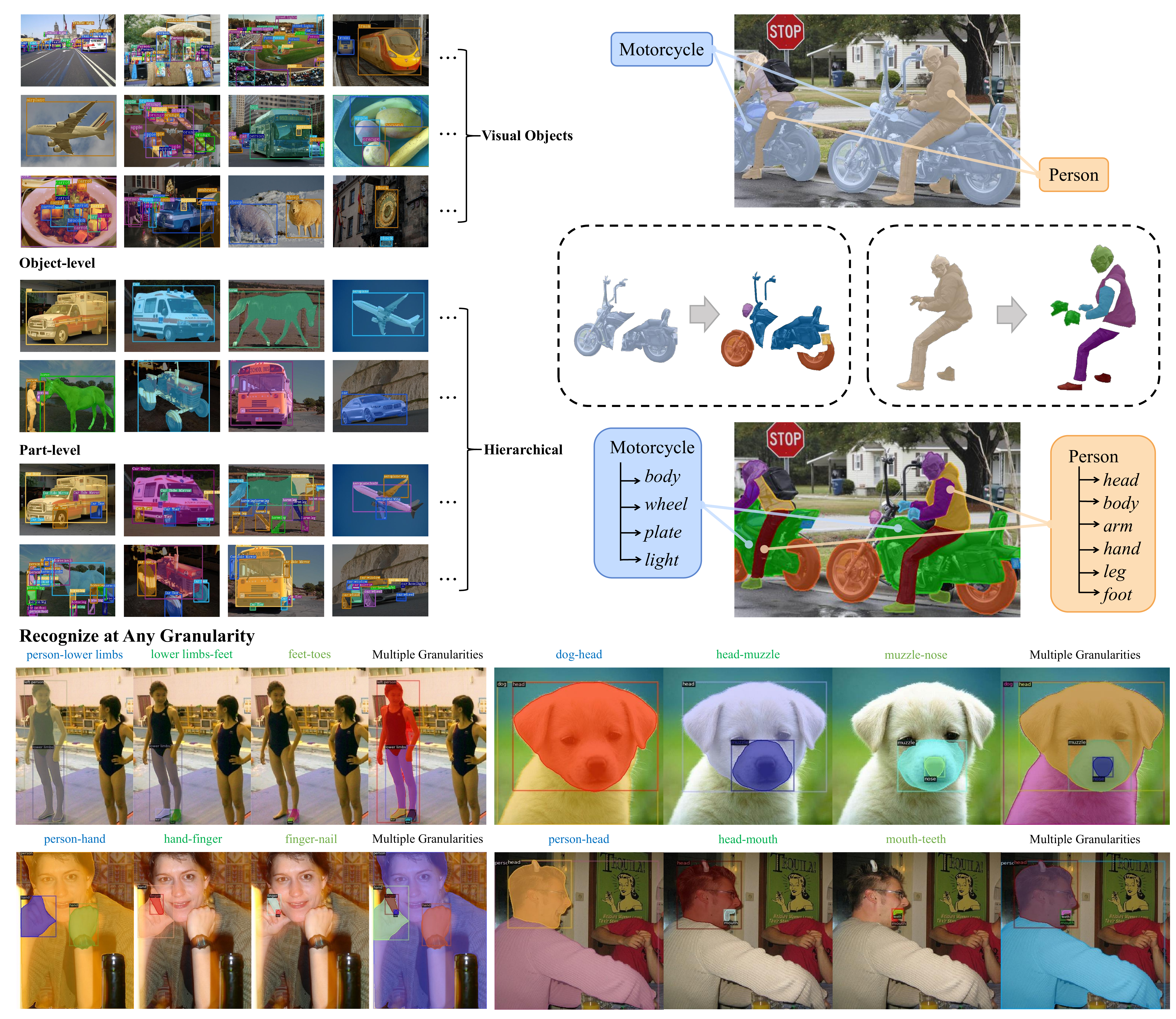}
\caption{
An illustrative example demonstrating image annotations at diverse granularities across multiple datasets. The annotations at hierarchical levels with corresponding relationships are depicted on the right side. Below is a visualization of our segmentation results at multiple granularities.
}
\label{fig:comparison}
\vspace{-2ex}
\end{figure*}

In recent years, foundation models have dominated the majority of tasks in the fields of Natural Language Processing~\cite{bert,gpt3,T5} and Computer Vision~\cite{CLIP,SAM,stable_diffusion,MAE, wu2023GLEE, florence2}.
CLIP family~\cite{CLIP,ALIGN,florence,fang2023eva,eva02} have made significant advancements in transfer learning and have demonstrated impressive zero-shot capabilities on vision-language tasks.
SAM~\cite{SAM} has revolutionized the development of segmentation tasks and is able to provide multi-level class-agnostic masks.
GLEE~\cite{wu2023GLEE} utilized diverse object-level data to develop general object representations, enabling detection, segmentation, tracking, grounding, and identification of objects in open-world scenarios. Their remarkable achievement can be attributed to the integration of extensive and diverse range of datasets.

Different from the vast quantity of object-level data, the scale of part-level data is relatively small, which turns out to be a major bottleneck hindering vision models from recognizing part-level instances. 
Thus, most vision models lack the hierarchical comprehension between objects and parts.
However, it is evident that the ability to recognize parts from objects is essential for various practical applications such as image editing\cite{kawar2023imagic, ling2021editgan, li2021partgan}, behavior analysis\cite{ng2022animal, yang2019parsing}, pose estimation\cite{Dong_2014_CVPR, yang2011articulated}, robotics manipulation\cite{brohan2022rt, nair2022r3m}, etc. 
Moreover, we observe that part-level information is able to help multi-modal Large Language Models (mLLMs) in achieving a more detailed understanding of image content.
Since part-level comprehension is a critical ability for foundation models to tackle a broader range of problems, it leads to a natural question: 
How could we break through data limitations to build a part-level vision foundation model?

To enable object foundation model with part-level cognitive ability, we emphasize that the model should achieve two key objectives: (1) \textbf{Hierarchical Comprehension}, the model is supposed to understand the intrinsic relationship between objects and parts, and extend this hierarchical connection to any novel object, (2) \textbf{Semantic Granularity}, the model should be capable of learning a universal feature representation, enabling it to recognize semantic instances at any granularity. Consequently,
we present a method to jointly detect and segment both objects and parts in a top-down manner. A lightweight Querying Transformer (Q-Former) is proposed to construct the hierarchical relationship between objects and parts. 
Specifically, it employs a set of universal parsing queries to interact with object queries, consequently generating multiple part-level queries that are capable of predicting corresponding semantic parts for each object. The Q-Former acts as a decomposer, which first recognizes individual objects in the images and subsequently parsing them into parts. Such model design is built upon the observation that various common objects often exhibit shared characteristics of parts. For example, cats, dogs, and dinosaurs all have parts such as torso, legs, and tails. In this way, two sets of query embeddings at different levels are generated, which are then used to predict object-level and part-level instances respectively. Through this approach, the relationship between objects and parts is established via the Q-Former design. Meanwhile, the hierarchical levels of objects and parts are distinguished, which is different from previous research\cite{chen2014detect, peize2023vlpart, wei2023ov, wang2024hierarchical, ramanathan2023paco} that consider parts as fine-grained objects. This paradigm enables vision models to better understand the features on different levels during training, thereby achieving improved performance.


Our complete solution, \methodNAME, for jointly detecting and segmenting instances at both object and part levels, makes it possible for vision models to achieve favorable outcomes on both object and part levels. Some previous research have devised specialized training paradigms to utilize abundant image-text pair data\cite{detclip, xdecoder} as well as grounding data\cite{GLIP, liu2023grounding, wu2023GLEE}, thereby enhancing the cognitive and generalization capabilities of the models. On the contrary, the quantity of part-level data is much smaller compared to object-level data. So far, the largest dataset incorporating the concepts of both objects and parts is the recently proposed PACO\cite{ramanathan2023paco} dataset. The scarcity of data has limited research on part-level recognition and restricted the generalization improvement of vision models. Although VLPart\cite{peize2023vlpart} has attempted to utilize pseudo-labeling schemes to generate part-level annotations for both object-level and image-level datasets, the quality of the pseudo-labels is relatively poor. Our innovative algorithm that parsing objects into their corresponding parts facilitates the transfer of generalization capability from objects to parts. Consequently, parts are generated from objects, which enables vision models to maintain generalization performance when predicting parts for novel objects without labeling extensive part-level data. 
To facilitate the training process of Q-Former, we standardize the annotation granularity across various part-level datasets and introduce a vast amount of object-level datasets, an intuitive display of the overall training data is shown in ~\cref{fig:comparison}.
Unlike VLPart, which exhibits unsatisfactory performance at object-level datasets after joint-training, our method demonstrates favorable outcomes at both object and part levels after joint-training. Moreover, it turns out that using object-level datasets is able to improve the performance of the model on part-level tasks, indicating a beneficial interaction between objects and parts.

Extensive experiments demonstrate that our method significantly improve the open-vocabulary part segmentation performance, concurrently ensuring a decent performance on object detection and segmentation. We verify its effectiveness on various popular datasets. To validate the generalization performance of our model in identifying various parts of novel objects, we conduct experiments on PartImageNet\cite{he2021partimagenet} and Pascal Part\cite{chen2014detect} datasets in cross-dataset and cross-category manners respectively. Our method exhibits strong transferability and generalization ability when adding extra object-level datasets during training. To evaluate the decomposition capability of our model, we conduct experiments on both ADE20K-Part and Pascal Part datasets follow OV-PARTS\cite{wei2023ov}. As a result, our approach significantly outperforms one-stage baselines of OV-PARTS, with an increase of 8.16\% and 2.07\% on harmonic mean IoU (hIoU) in ADE20K-Part-234 and Pascal-Part-116 respectively. 
Additionally, by incorporating a large amount of object-level data for joint-training, our method establishes generic hierarchical relationships and breaks through the limitations of scarce part-level data, achieving state-of-the-art performance across various part-level tasks.


In conclusion, our main contributions can be summarized as follows:
\begin{enumerate}
\item We construct the hierarchical relationship between objects and parts via the Q-Former, facilitating part segmentation to acquire advantages from various object-level datasets.
\item We propose a unified pipeline for hierarchical detection and segmentation, where we first recognize objects and then parsing them into corresponding semantic parts. This algorithm enables us to jointly detect and segment both object-level and part-level instances.
\item We standardize the annotation granularity across various part-level datasets by incorporating corresponding object-level annotations, complementing the hierarchical correspondences for current part-level datasets, promoting the development of vision foundation models.
\end{enumerate}
\section{Related Work}
\label{sec:Related}


\subsection{Visual Foundation Models and Generalist Models}
Visual foundation models and generalist models are considered as a milestone in the development of the intelligent vision system. 
For instance, multi-modal visual foundation models~\cite{CLIP,ALIGN, florence, beit3,flamingo} have significantly advanced efficient transfer learning and exhibit impressive zero-shot capabilities on vision-language tasks by using contrastive learning with large-scale image-text pairs. Generative foundation models~\cite{DALLE,DALLE2,vqgan, stable_diffusion} are trained on vast collections of images and captions, empowering them to generate image content conditioned on textual prompts. Self-supervised foundation models~\cite{sslDINO, MAE, fang2023eva,eva02} have learned general visual representations from large-scale image datasets, enhancing their ability to transfer to downstream tasks. However, the image-level features learned by these foundation models are not well-suited for direct application to dense prediction tasks that involve precise object and part localization.

Transformer-based generalist methods~\cite{Uni-perceiver, OFA, unified-io, Pix2Seqv2, unitab} adopt a sequence generation pipeline to unify the output of text and spatial coordinates. However, they mainly focus on image-level comprehension, which results in relatively weak localization capabilities.
Works such as UNINEXT, etc.\cite{UNINEXT,Uni-perceiverv2,unicorn}, 
built upon strong detectors\cite{maskdino, deformableDETR}, demonstrating a strong localization capability across multiple datasets. But they fail to exhibit zero-shot transfer ability and generalization capability due to their closed-set training paradigm. 
Some works about open-vocabulary detection (OVD)\cite{zareian2021open, zhong2022regionclip, GLIP, lin2022learning, yao2023detclipv2, codet, lin2024generative, ma2024groma} have explored zero-shot generalization capabilities on novel categories.
X-Decoder\cite{xdecoder} and SEEM\cite{SEEM} have developed a versatile decoding architecture that are able to generate accurate pixel-level segmentation predictions. 
GLEE\cite{wu2023GLEE} addresses various object-level tasks through a unified architecture and training paradigm. However, current generalist models and foundation models are trained mainly on image-level and object-level datasets, thus their ability to extract more fine-grained information is limited, making it difficult for them to recognize corresponding semantic parts of any object. Our work focuses on empowering hierarchical cognitive capability for vision foundation models, thereby further advancing the development of comprehensive visual systems.

\subsection{Part Segmentation}
The growing interest in achieving a more fine-grained understanding of objects has sparked a surge in research focused on part level recognition. Some pioneering studies have introduced datasets with part-level annotations, concentrating on objects of some specific categories such as human body parts\cite{gong2017look,li2017multiple,yang2019parsing}, animal body parts\cite{WahCUB_200_2011} and vehicle components\cite{reddy2018carfusion}. More general part annotations for common objects such as Pascal-Part\cite{chen2014detect}, PartImageNet\cite{he2021partimagenet}, ADE20K\cite{zhou2018semantic}, CityscapesPanoptic-Parts\cite{meletis2020cityscapes} and more recent PACO\cite{ramanathan2023paco} were then proposed to promote more in-depth research in the field of parts. Most of the previous works\cite{degeus2021panopticparts, li2022panoptic, zhou2021differentiable,michieli2020gmnet} were conducted based on a closed-set configuration, thus only capable of detecting and segmenting closed-set objects and parts. Recently, VLPart\cite{peize2023vlpart} present a pipeline for detecting and segmenting both open-vocabulary objects and their corresponding part regions, while OV-PARTS\cite{wei2023ov} utilize adapters to transfer the generalization abilities of CLIP into open-vocabulary part segmentation task. However, due to the limited quantity of data, the generalization capability of previous models\cite{gong2017look,li2017multiple,yang2019parsing,peize2023vlpart,wei2023ov,tang2022request,pan2023towards,wang2024hierarchical} still relies heavily on the training datasets. Furthermore, in prior works, both objects and parts are treated equally, they consider part as a special type of object. On the contrary, we distinguish them by considering parts as integral components subordinate to objects and generate parts from corresponding objects in a top-down manner. 
Our work is aimed at building hierarchical relationships while unifying the training paradigm for object-level and part-level data.
By incorporating a large amount of object-level data, the hierarchical relationships can be extended to any object, enabling our method to recognize a rich variety of parts. 


\subsection{Hierarchical Learning of Objects and Parts}
Learning objects through parts has been a long-standing research topic as part annotations provide more detailed semantic information of objects. Morabia et al.\cite{morabia2020attention} first introduced a pipeline employing an attention mechanism for simultaneous detection of both objects and parts. Deepflux\cite{deepflux} designed an image context flux representation which enables better object parts interaction for skeleton detection. Leopart\cite{ziegler2022self} demonstrated that learning object parts can provide spatially diverse representation which facilitates self-supervised semantic segmentation. Wang et al.\cite{wang2015joint} proposed a method to predict both parts and objects simultaneously on Pascal-Part dataset\cite{chen2014detect}. Recent studies such as SAM\cite{SAM} and Semantic-SAM\cite{li2023semantic} have studied on class-agnostic multi-granularity interactive segmentation task. However, they have not explored the relationship between objects and their corresponding semantic parts. Recently, Compositor\cite{He_2023_CVPR} designed a bottom-up pipeline to predict parts and then cluster them into objects, while AIMS\cite{qi2024aims} utilized an independent relation decoder to construct the hierarchical association between objects and parts. Different from these works, our approach introduces a Querying Transformer to model the hierarchical relationship, allowing our model to parse any object into its corresponding parts.
\section{Method}
\label{sec:Method}

\begin{figure*}[tb]
\centering
\includegraphics[width=0.9 \linewidth]{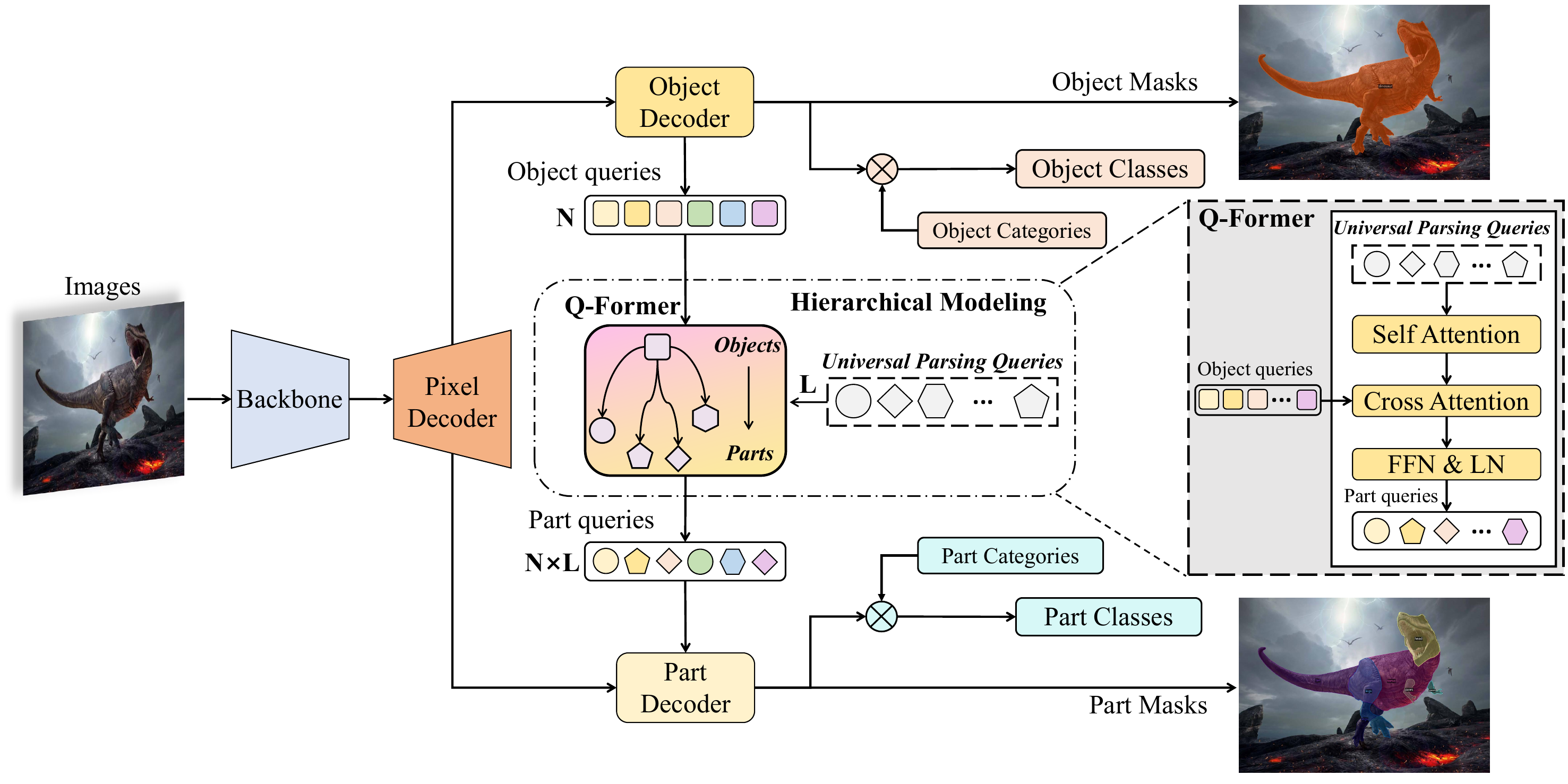}
\caption{
\textbf{Framework of \methodNAME.}
The Q-Former takes each object query as input and output the corresponding part queries. These queries are then fed into the object decoder and the part decoder respectively to generate hierarchical predictions. 
}
\label{fig:framework}
\vspace{-2ex}
\end{figure*}


\subsection{Overall Framework}
\label{sec:Overall Framework}
Following \cite{wu2023GLEE, wang2024hierarchical}, we propose \methodNAME, which comprises of an image encoder, a Q-Former, two independent decoders and a text encoder, as shown in \cref{fig:framework}. 


Given an input image $I \in \mathcal{R}^{H\times W \times 3}$, the backbone and the pixel decoder first extract multi-scale image features $F_s \in \mathcal{R}^{\frac{H}{2^s} \times \frac{W}{2^s} \times C}$ and $s=\{2,3,4,5\}$ with backbones such as ResNet\cite{resnet} or Swin Transformer\cite{SwinTransformer}. 
Then we feed them into the object decoder, where the object-level query embeddings $q_{obj} \in \mathcal{R}^{N\times C}$ are generated in a two-stage process. These object queries are utilized to perform object-level classification, detection as well as segmentation tasks through three independent prediction heads. Besides, the object queries $q_{obj}$ are fed into the Q-Former simultaneously, where $L$ learnable universal parsing queries are initialized to interact with object queries. 
It takes object queries as input and generate part-level queries $q_{part} \in \mathcal{R}^{N\cdot L \times C}$ which are then passed into the part decoder to yield part-level predictions (detailed in \cref{sec:Q-Former}).
{To enhance the semantic-awareness, an early fusion module is adopted before Transformer encoder following \cite{UNINEXT}, which takes image feature from backbone and text embedding as input and perform bi-directional cross-attention between them.}
In line with previous segmentation models\cite{mask2former, maskdino, ViTDet}, a pixel embedding map $M_p \in \mathcal{R}^{\frac{H}{4} \times \frac{W}{4} \times C} $ at 1/4 resolution is constructed by upsampling and integrating multi-scale feature maps from the backbone and the pixel decoder. Eventually, we dot product each object query or part query with the pixel embedding map to derive an output mask $m \in \mathcal{R}^{\frac{H}{4} \times \frac{W}{4}}$:
\begin{align}
    \vspace{-2ex}
    m = FFN(q_l) \otimes M_p, \quad l\in \{obj, part\}
    \label{eq:1}
    \vspace{-2ex}
\end{align}
where FFN is comprised of 3 layers feed forward network with ReLU activation functions and linear layers.

\subsection{Parsing Objects into Parts}
\label{sec:Q-Former}
We propose a Q-Former to establish the hierarchical relationship between objects and parts.
As various common objects tend to manifest shared attributes in their constituent parts, for example, both lizards and birds exhibit similar components, such as heads and torsos. Thus, we initialize a set of query embeddings in the Q-Former to parse any object into semantic parts. 
We denote these universal parsing query embeddings as $q_{parse} \in \mathcal{R}^{L \times C}$, where $L$ represents the number of the parsing queries. As shown in \cref{fig:framework}, the Q-Former is comprised of $M$ cascaded attention modules, each module includes a self-attention layer, a cross-attention layer, and a feed forward network. The universal parsing queries are first fed into the self-attention layer and then perform cross-attention with the object queries. Note that every object query is interacted with all universal parsing queries. Hence, assume $N$ object queries($q_{obj} \in \mathcal{R}^{N\times C}$) are generated from the object decoder, and $L$ universal parsing queries $q_{parse} \in \mathcal{R}^{L \times C}$ are initialized in the Q-Former, we obtain $N\cdot L$ part-level queries which can be denoted as $q_{part} \in \mathcal{R}^{N\cdot L\times C}$. We refer to this process as:
\begin{align}
    \vspace{-2ex}
    q_{part} = Q\mbox{-}Former(q_{parse}; q_{obj})
    \label{eq:2}
    \vspace{-2ex}
\end{align}

Our proposed Q-Former functions as a decomposer, extracting and representing parts from object queries. 
Hence, by training jointly on extensive object-level datasets and limited hierarchical datasets which contain object-part correspondences, our Q-Former obtains strong generalization ability to parse any novel object into its corresponding parts.
\subsection{Unified Training Paradigm for Objects and Parts}
\label{sec:Training Paradigm}
\begin{figure*}[tb]
\centering
\includegraphics[width=0.99 \linewidth]{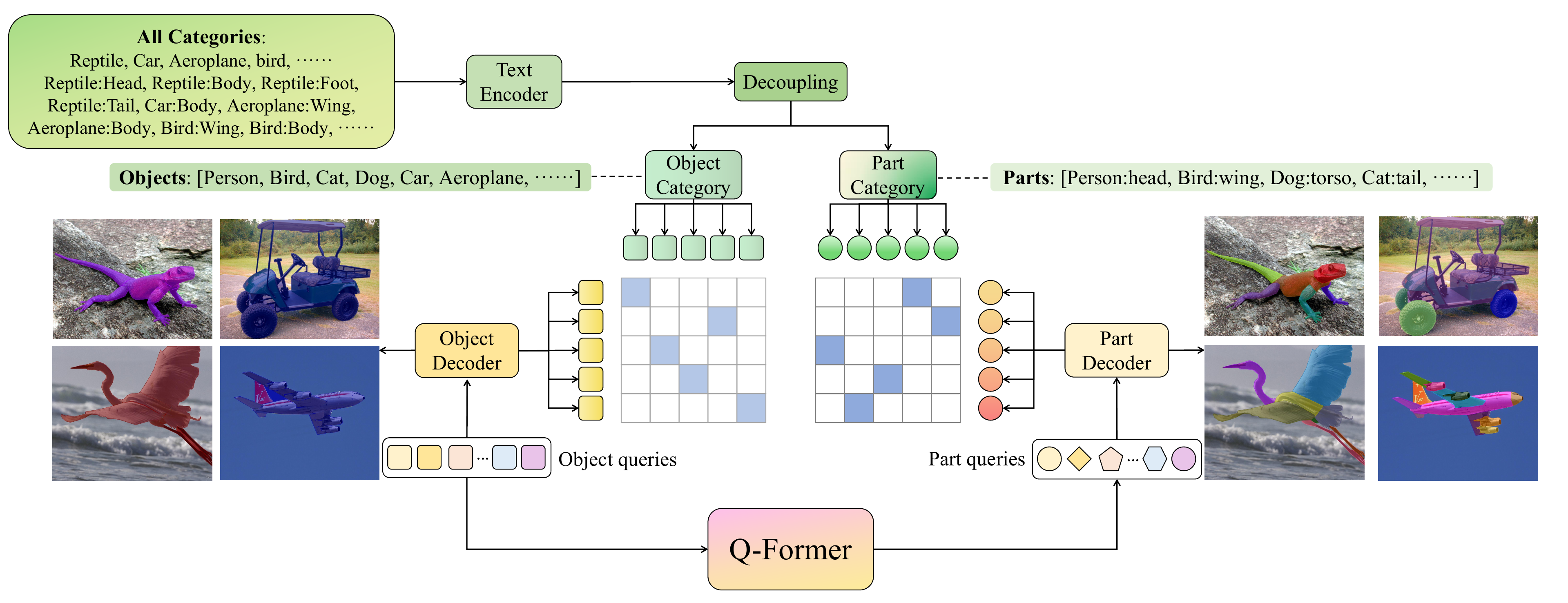}
\caption{
\textbf{Matching mechanisms of \methodNAME.}
Two separate forward passes are performed on the same image to obtain hierarchical segmentation results.
}
\label{fig:matching_mechanism}
\vspace{-2ex}
\end{figure*}
Since the Q-Former requires hierarchical data to learn how to parse objects into parts, we enrich part-level data with corresponding object-level annotations. Details are provided in the appendix.

Since the annotation granularity across part-level datasets is standardized, our model can first learn the characteristics of objects and then acquire the ability to parse any object into its semantic parts. 
To facilitate open-vocabulary detection and segmentation, we substitute the similarity scores between the instance embeddings and the text embeddings for the original class head. Given $K$ object-level and part-level input categories as separate sentences, we feed them into the text encoder and utilize the average of each individual sentence tokens as the output text embedding $T_l$ for each category. Then the similarity scores $S_l \in \mathcal{R}^{N\times K}$ are calculated through a dot product operation between the hierarchical instance embeddings $q_l \in \mathcal{R}^{N\times C}$ from detector and the text embeddings $T_l \in \mathcal{R}^{K\times D}$ from text encoder, which can be denoted as:
\begin{align}
    \vspace{-2ex}
    S_l = q_l \cdot W_{proj} \otimes T_l, \quad l\in \{obj, part\}
    \label{eq:4}
    \vspace{-2ex}
\end{align}
where $W_{proj} \in \mathcal{R}^{C \times D}$ is a trainable projection weight for fine-tuning text embedding space especially for part-level descriptions. Following \cite{wang2024hierarchical, li2023semantic}, we perform Hungarian matching of objects and parts individually, where object-level predictions are only matched with object-level targets, and the same applies to the part-level output, as shown in \cref{fig:matching_mechanism}. 

We then introduce a constraint loss to ensure the part-level predictions to be the component of the objects. We denote this novel loss function as \textbf{restriction loss} $L_{res}$. Due to memory limitations, we only calculate our restriction loss on the predicted bounding boxes between different levels, while leaving the predicted masks unconstrained. Our restriction loss can be calculated as follow:
\begin{align}
    \vspace{-2ex}
    L_{res} = \sum_{i}^L (1 - \frac{\lvert S_{obj}\cap S_{part}^i\rvert}{S_{part}^i})
    \label{eq:6}
    \vspace{-2ex}
\end{align}
where $S_{obj}$ represents the area of the object-level bounding box prediction, and $S_{part}^i$ stands for the area of the $i-th$ part-level bounding box prediction. Note that each object query can generate $L$ part queries through Q-Former. This loss function is only applied to the matched predictions in part-level datasets, thereby strengthening the mutual correspondence between different hierarchies.

\methodNAME is trained with a linear combination of losses for object-level tasks and part-level tasks, which can be formulated as:
\begin{align}
    \vspace{-2ex}
    L = \lambda_{1}(L_{cls}^{obj} + L_{cls}^{part}) + \lambda_{2}(L_{box}^{obj} + L_{box}^{part}) + \lambda_{3}(L_{mask}^{obj} + L_{mask}^{part}) + \lambda_{4}L_{res}
    \label{eq:7}
    \vspace{-2ex}
\end{align}
where $L_{cls}^l$, $L_{box}^l$, $L_{mask}^l$ are the classification, box, and mask loss at different levels ($l\in \{obj, part\}$), while $L_{res}$ is the restriction loss, and $\lambda$ are their corresponding weights. 
We apply Focal Loss\cite{focalloss} as the classification loss on the similarity scores $S_l$ to align the text concepts with instance features. A combination of L1 loss and generalized IoU loss\cite{giou} is utilized for box predictions, while we employ both Dice Loss\cite{diceloss} and Focal Loss to calculate mask loss. 
We follow MaskDINO to set our hyperparameters to $\lambda_1=4, \lambda_2=2, \lambda_3=5, \lambda_4=5$.
Based on the above designs, \methodNAME is able to leverage both object-level data and part-level data thus obtaining a strong generalization capability. 
\section{Experiments}
\subsection{Experimental Setup}

We conduct comprehensive experiments to exhibit the effectiveness of \methodNAME across a wide range of object-level and part-level tasks.

\setlength{\tabcolsep}{4pt}
\begin{table}[t]
\caption{The data statistics for joint-training in \methodNAME.}
\begin{center}
\vspace{-1em}
\resizebox{0.9\columnwidth}{!}{
\begin{tabular}{lllccccccc}

\hline\noalign{\smallskip}
\multirow{2}{*}{Type}&\multirow{2}{*}{Datasets}&\multirow{2}{*}{Images}&\multicolumn{2}{c}{Semantic Concept}&\multicolumn{3}{c}{Annotations} \\
\cmidrule(lr){4-5} \cmidrule(lr){6-8}
 & & & Object & Part & Semantic & Box & Mask \\
\hline
\multirow{12}{*}{Object-Level}
 & Object365\cite{objects365} & 1.8M & 365 & \usym{2613} & Category & \checkmark & \usym{2613}\\
 & OpenImages\cite{OpenImages} & 1.7M & 601 & \usym{2613} & Category & \checkmark & \usym{2613} \\
 & COCO\cite{coco} & 118K & 80 & \usym{2613} & Category & \checkmark & \checkmark\\
 & LVIS\cite{lvis} & 100K & 1203 & \usym{2613} & Category & \checkmark & \checkmark \\
 & BDD\cite{bdd100k} & 70K & \checkmark & \usym{2613} & Category & \checkmark & \checkmark \\
 & UVO\cite{UVO} & 70K & \checkmark & \usym{2613} & \usym{2613} & \checkmark & \checkmark \\
 & YTVIS19\cite{ytvis2019} & 62K & \checkmark & \usym{2613} & Category & \checkmark & \checkmark \\
 & YTVIS21\cite{ytvis21dataset} & 90K & \checkmark & \usym{2613} & Category & \checkmark & \checkmark \\
 & OVIS\cite{OVIS} & 42K & \checkmark & \usym{2613} & Category & \checkmark & \checkmark \\
 & RefCOCO\cite{RefCOCOandplus} & 17K & \checkmark & \usym{2613} & Description & \checkmark & \checkmark \\
& RefCOCOg\cite{RefCOCOg-umd} & 22K & \checkmark & \usym{2613} & Description & \checkmark & \checkmark \\
& RefCOCO+\cite{RefCOCOandplus} & 17K & \checkmark & \usym{2613} & Description & \checkmark & \checkmark \\
\hline
\multirow{6}{*}{Hierarchical}
 & PACO\cite{ramanathan2023paco} & 77K & 75 & 456 & Category & \checkmark & \checkmark\\
 & Pascal Part\cite{chen2014detect} & 5K & 20 & 93 & Category & \checkmark & \checkmark\\
 & PartImageNet\cite{he2021partimagenet} & 16K & 11 & 40 & Category & \checkmark & \checkmark\\
 & ADE20K-Part\cite{wei2023ov} & 8K & 44 & 234 & Category & \checkmark & \checkmark\\
 & Visual Genome\cite{visualgenome} & 108K & \checkmark & \checkmark & Description & \checkmark & \usym{2613} \\
 & SA-1B\cite{SAM} & 110K &\checkmark & \checkmark & \usym{2613} & \checkmark & \checkmark\\
\bottomrule
\end{tabular}}
\end{center}
\label{table:data_statistics}
\vspace{-2em}
\end{table}
\setlength{\tabcolsep}{1.4pt}

\textbf{Data Unification}. 
We utilize object-level datasets such as COCO\cite{coco}, LVIS\cite{lvis}, Object365\cite{objects365},
OpenImages\cite{OpenImages}, Visual Genome\cite{visualgenome} and RefCOCO series \cite{RefCOCOandplus, RefCOCOg-umd}, etc, while using part-level datasets PACO\cite{ramanathan2023paco}, PartImageNet\cite{he2021partimagenet}, Pascal Part\cite{chen2014detect}, ADE20K-Part\cite{wei2023ov} and SA-1B\cite{SAM} with varying annotation granularity for joint-training. 
For Visual Genome and SA-1B, we categorize their corresponding part-level annotations based on semantic and mask overlap relationships to construct hierarchical data versions. For part-level data, we integrate the original part-level annotations with corresponding object-level annotations according to their associated object-level dataset. The details of these dataset preprocessing steps are left in the appendix. 
The datasets used for joint-training and their statistical characteristics are shown in \cref{table:data_statistics}. 

\textbf{Implementation Details}. In our experiments, we utilize ResNet-50\cite{resnet} and Swin-Large\cite{SwinTransformer} as the vision encoder. Following MaskDINO\cite{maskdino}, we adopt deformable transformer in the decoder, and use 300 object queries while setting the number of parsing queries $L$ to be 10. The $M$ of Q-Former is set to 6.
We select the top 50 object queries based on the similarity scores and input them into the Q-Former, ultimately yielding 500 part queries.
We use both query denoising and hybrid matching strategies to facilitate convergence and enhance performance. We conduct experiments on part-level datasets following the methodologies of VLPart\cite{peize2023vlpart} and OV-PARTS\cite{wei2023ov} in order to evaluate the generalization performance and the ability to parse novel objects of our model. For joint-training, we train \methodNAME based on the weights of GLEE\cite{wu2023GLEE}, continuing training on 32 A100 GPUs. The settings for the part-level zero-shot experiments are described separately in each section.

\subsection{Zero-shot Part Segmentation Results}
\setlength{\tabcolsep}{4pt}
\begin{table}[t]
\caption{Cross-dataset generalization performance compared with VLPart. The evaluation metric is $mAP_{mask}$ on the validation set of PartImageNet. All models utilize ResNet-50 as backbone and use the text embeddings of the category names as the classifier. \sethlcolor{gray!20}\hl{PartImageNet} denotes the fully-supervised method reported for comparison.}
\begin{center}
\vspace{-1em}
\resizebox{0.7\columnwidth}{!}{
\begin{tabular}{llccccccc}
\hline\noalign{\smallskip}
\multirow{2}{*}{Method}   &\multirow{2}{*}{Datasets}  &\multirow{2}{*}{All} &\multicolumn{4}{c}{\textit{quadruped}} \\
\cmidrule(lr){4-7}
 & & (40) &\textit{head} &\textit{body} &\textit{foot} &\textit{tail} \\
\noalign{\smallskip}
\hline
\noalign{\smallskip}
\multirow{5}{*}{VLPart\cite{peize2023vlpart}}
 &Pascal Part & 4.5 & 17.4 & 0.1 & 0.0 & 2.9  \\
 &+ IN-S11 label & 5.4 & 23.6 & 3.4 & 0.8 & 1.2  \\
 &+ Parsed IN-S11 & 7.8 & 35.0 & 15.2 & 3.5 & 8.9 \\
 &\textcolor{darkgreen}{\emph{vs. baseline}}&\textcolor{darkgreen}{+3.3} &\textcolor{darkgreen}{+17.6}&\textcolor{darkgreen}{+15.1}&\textcolor{darkgreen}{+3.5}&\textcolor{darkgreen}{+6.0}\\
\cline{2-7}
&\cellcolor{gray!20}\textcolor{black}{PartImageNet}
&\cellcolor{gray!20}\textcolor{black}{29.7}
&\cellcolor{gray!20}\textcolor{black}{57.3}
&\cellcolor{gray!20}\textcolor{black}{25.8}
&\cellcolor{gray!20}\textcolor{black}{22.9}
&\cellcolor{gray!20}\textcolor{black}{22.9}
\\
\hline
\multirow{4}{*}{\methodNAME}
 &Pascal Part & 9.9 &23.6 &4.5 &1.3 &4.6  \\
 &+ Parsed IN-S11 & 14.9 & 55.3 & 27.2 & 7.0 & 23.6 \\
 & \textcolor{darkgreen}{\emph{vs. baseline}}&\textcolor{darkgreen}{+5.0}&\textcolor{darkgreen}{+31.7}&\textcolor{darkgreen}{+22.7}&\textcolor{darkgreen}{+5.7}&\textcolor{darkgreen}{+19.0}\\
 \cline{2-7}
&\cellcolor{gray!20}\textcolor{black}{PartImageNet}
&\cellcolor{gray!20}\textcolor{black}{40.2}
&\cellcolor{gray!20}\textcolor{black}{67.0}
&\cellcolor{gray!20}\textcolor{black}{37.6}
&\cellcolor{gray!20}\textcolor{black}{36.5}
&\cellcolor{gray!20}\textcolor{black}{40.7}
\\
\hline
\end{tabular}}
\end{center}
\label{table:cross_dataset_results}
\vspace{-2em}
\end{table}
\setlength{\tabcolsep}{1.4pt}

\textbf{1) Cross-dataset Part Segmentation on PartImageNet.}
We follow VLPart\cite{peize2023vlpart} to conduct experiments on cross-dataset generalization performance by directly evaluating on PartImageNet\cite{he2021partimagenet} validation set. We report the metrics of all (40) part categories and the detailed metrics of \textit{quadruped} are also provided. The baseline approach only utilize Pascal Part as the training set and directly perform evaluation on PartImageNet in a zero-shot manner. Note that \textbf{IN-S11 label} represents adding image-level classification data for training in order to improve performance. Meanwhile, \textbf{Parsed IN-S11} stands for training with the pseudo-labels generated from the parsing pipeline proposed by VLPart. 
However, both of these methods expose the model to categories and images from the PartImageNet dataset. 
We first perform our training process exclusively on the Pascal Part dataset to verify our zero-shot capabilities, and then we incorporate pseudo-labels to assess the ability of our model to utilize low-quality annotations.


Given that Pascal Part does not provide semantic labels for categories like \textit{quadruped} in PartImageNet, the model needs to generalize 
from annotated parts of \textit{dog}, \textit{cat}, etc. in Pascal Part to parts of \textit{quadruped} in PartImageNet. As shown in \cref{table:cross_dataset_results}, our model significantly outperform VLPart when only training on Pascal Part, even surpassing the model trained with \textbf{Parsed IN-S11}. 
After incorporating pseudo-labeled data into training, our model shows higher performance gains, indicating better utilization of low-quality data.
This result illustrates the importance of hierarchical modeling, which enables our model to recognize and parse novel objects into their corresponding parts based on the generalization capability brought by CLIP.

\setlength{\tabcolsep}{4pt}
\begin{table}[t]
\caption{Cross-category generalization performance compared with VLPart. The evaluation metric is $mAP_{mask}$ on the validation set of PascalPart but report $AP50$ specifically for dog parts. All models utilize ResNet-50 as backbone and use the text embeddings of the category names as the classifier. Base part
represents the base split from Pascal Part. VOC object is added to the training process to improve the cognitive ability of the model thus reach a better performance. \sethlcolor{gray!20}\hl{Pascal Part} denotes the fully-supervised method reported for comparison.}
\begin{center}
\vspace{-1em}
\resizebox{0.98\columnwidth}{!}{
\begin{tabular}{llccccccccc}
\hline\noalign{\smallskip}
\multirow{2}{*}{Method} & \multirow{2}{*}{Datasets} & \multirow{2}{*}{All AP} & \multirow{2}{*}{BaseAP} &\multirow{2}{*}{NovelAP} &\multicolumn{5}{c}{\textit{dog}} & \multirow{2}{*}{NovelAP}\\
\cmidrule(lr){6-10}
 & & (93) & (77) & (16) & \textit{head} &\textit{torso} &\textit{leg} &\textit{paw} & \textit{tail} & Increment \\
\noalign{\smallskip}
\hline
\noalign{\smallskip}
\multirow{5}{*}{VLPart\cite{peize2023vlpart}}
 & Base Part & 15.0 & 17.8 & 1.5 & 6.1 & 7.9 & 2.9 & 13.8 & 3.2 & -  \\
 & + VOC object & 16.8 & 19.9 & 2.1 & 29.9 & 22.6 & 3.2 & 12.4 & 2.1 & \textcolor{darkgreen}{\textit{+0.6}}  \\
 & + IN-S20 label & 17.4	& 20.8 & 1.1 & 12.8 & 17.8 & 2.0 & 5.9 & 0.9 & \textcolor{darkgreen}{\textit{-0.4}}\\
 & + Parsed IN-S20 & 18.4 & 21.3 & 4.2 &	28.7 & 34.8 & 17.2& 5.7 &14.3 & \textcolor{darkgreen}{\textit{+2.7}}\\
\cline{2-11}
& \cellcolor{gray!20}\textcolor{black}{Pascal Part}
& \cellcolor{gray!20}\textcolor{black}{19.4}
& \cellcolor{gray!20}\textcolor{black}{18.8}
& \cellcolor{gray!20}\textcolor{black}{22.4}
& \cellcolor{gray!20}\textcolor{black}{88.0}
& \cellcolor{gray!20}\textcolor{black}{49.6} 
& \cellcolor{gray!20}\textcolor{black}{38.3} 
& \cellcolor{gray!20}\textcolor{black}{48.9} 
& \cellcolor{gray!20}\textcolor{black}{25.8} 
& \cellcolor{gray!20}\textcolor{black}{-} 
\\
\hline
\multirow{4}{*}{\methodNAME}
 & Base Part & 25.6 & 30.5 & 2.1 & 12.6 & 15.6 & 8.2 & 5.2 & 6.2 & -  \\
 & + VOC object & 26.9 & 31.2 & 5.8 & 46.5 & 35.0 & 27.0 & 14.7 & 15.1 & \textcolor{darkgreen}{\textit{+3.7}}  \\
 & + Parsed IN-S20 & 26.6 & 28.9 & 15.5 & 80.3 & 57.3 & 36.7 & 17.0 & 37.4 & \textcolor{darkgreen}{\textit{+9.7}}  \\
 \cline{2-11}
& \cellcolor{gray!20}\textcolor{black}{Pascal Part} 
& \cellcolor{gray!20}\textcolor{black}{35.5} 
& \cellcolor{gray!20}\textcolor{black}{34.6} 
& \cellcolor{gray!20}\textcolor{black}{39.9} 
& \cellcolor{gray!20}\textcolor{black}{95.9} 
& \cellcolor{gray!20}\textcolor{black}{88.5} 
& \cellcolor{gray!20}\textcolor{black}{75.0} 
& \cellcolor{gray!20}\textcolor{black}{76.7} 
& \cellcolor{gray!20}\textcolor{black}{72.9} 
& \cellcolor{gray!20}\textcolor{black}{-} 
\\
\hline
\end{tabular}}
\end{center}
\label{table:cross_category_results}
\vspace{-2em}
\end{table}
\setlength{\tabcolsep}{1.4pt}

\textbf{2) Cross-category Part Segmentation on Pascal Part.}
We follow the evaluation setting proposed by VLPart to assess the cross-category generalization performance of our model on the Pascal Part dataset. A total of 93 part categories are divided into 77 base part categories and 16 novel part categories. \cref{table:cross_category_results} presents the evaluation results for all (93), base (77), and novel (16) parts. The model is trained only on the base categories, and is directly evaluated on the entire datasets. 
Note that \textbf{IN-S20 label} represents adding image-level classification data and \textbf{Parsed IN-S20} is on behalf of he pseudo-labels generated by VLPart\cite{peize2023vlpart} on ImageNet\cite{imagenet}.
We further introduce a metric called \textbf{NovelAP Increment} on top of VLPart to assess the improvement of our model when adding extra object datasets into the training process. It is calculated by subtracting the baseline Novel AP from the Novel AP achieved after incorporating extra datasets. 
The results shown in \cref{table:cross_category_results} demonstrate that our method surpasses the performance of VLPart by a large margin. By comparing the NovelAP Increment, we observe that our method achieves a greater increment after incorporating extra object dataset.
Since the VOC dataset includes object categories corresponding to novel parts, the hierarchical relationships of the Q-Former can be extended to novel part categories, resulting in a higher NovelAP Increment.

\setlength{\tabcolsep}{4pt}
\begin{table}[t]
\caption{Generalized zero-shot part segmentation performance on ADE-Part-234 and Pascal-Part-116 compared with baselines proposed by OV-PARTS.}
\begin{center}
\vspace{-2em}
\resizebox{0.98\columnwidth}{!}{
\begin{tabular}{clllcccccc}
\hline
\noalign{\smallskip}
\multirow{3}{*}{Method} & \multirow{3}{*}{Model} & \multirow{3}{*}{Backbone} & \multirow{3}{*}{Finetuning} & \multicolumn{6}{c}{Oracle-Obj}  \\
\cmidrule(lr){5-10}
 & & & & \multicolumn{3}{c}{ADE-Part-234} & \multicolumn{3}{c}{Pascal-Part-116} \\
\cmidrule(lr){5-7} \cmidrule(lr){8-10}
 & & & & Seen & Unseen & Harmonic & Seen & Unseen & Harmonic \\
\hline
Fully & \multirow{2}{*}{Mask2Former} & \multirow{2}{*}{ResNet-50} & \multirow{2}{*}{\usym{2613}} & \multirow{2}{*}{46.25} & \multirow{2}{*}{47.86} & \multirow{2}{*}{-} & \multirow{2}{*}{55.28} & \multirow{2}{*}{52.14} & \multirow{2}{*}{-} \\
Supervised & & & & & & & & \\
\hline
\multirow{3}{*}{Two-Stage} & 
\multirow{3}{*}{ZSseg$+$} & ResNet-50 & CPTCoOp & 43.19 & 27.84 & 33.85 & 55.33 & 19.17 & 28.48 \\
& & ResNet-50 & CPTCoCoOp & 39.67 & 25.15 & 30.78 & 54.43 & 19.04 & 28.21 \\
& & ResNet-101c & CPTCoOp & 43.41 & 25.70 & 32.28 & 57.88 & 21.93 & 31.81 \\
\hline
\multirow{5}{*}{One-Stage}
&\multirow{2}{*}{CATSeg} & ResNet-101\&ViT-B/16 & \usym{2613} & 11.49 & 8.56 & 9.81 & 14.89 & 10.29 & 12.17 \\
& & ResNet-101\&ViT-B/16 & B$+$D & 31.40 & 25.77 & 28.31 & 43.97 & 26.11 & 32.76 \\
\cmidrule(lr){2-10}
&\multirow{2}{*}{CLIPSeg} & ViT-B/16 & \usym{2613} & 15.27 & 18.01 & 16.53 & 22.33 & 19.73 & 20.95 \\
& & ViT-B/16 & VA$+$L$+$F$+$D & 38.96 & 29.65 & 33.67 & 48.68 & 27.37 & 35.04 \\
\cline{2-10}
& \cellcolor{gray!20}\methodNAME 
& \cellcolor{gray!20}ResNet-50 
& \cellcolor{gray!20}\usym{2613} 
& \cellcolor{gray!20}51.29 
& \cellcolor{gray!20}35.33 
& \cellcolor{gray!20}41.83 
& \cellcolor{gray!20}57.43 
& \cellcolor{gray!20}27.41 
& \cellcolor{gray!20}37.11 
\\
\hline
\end{tabular}}
\end{center}
\label{table:generalized_part_segmentation}
\vspace{-2em}
\end{table}
\setlength{\tabcolsep}{1.4pt}
\textbf{3) Generalized Zero-shot Part Segmentation.}
We adopt the \textbf{Oracle-Obj setting} proposed by OV-PARTS\cite{wei2023ov} to conduct experiments on ADE-Part-234 and Pascal-Part-116 datastes. This setting assumes that the ground-truth masks and categories of object-level instances are known during the inference process, aiming to evaluate the capability of the model to parse any novel object. All categories in the datasets are divided into a base set and a novel set, and the training process is performed only on the base set, while we evaluate the performance of the model on all categories. 
As shown in \cref{table:generalized_part_segmentation}, our model achieves a superior performance on both datasets, which indicates the importance of hierarchical modeling. 
The establishment of hierarchical relationships between objects and parts enables our model to extend  to novel objects, thereby effectively parsing them into corresponding semantic parts. Consequently, our model exhibits outstanding performance across both datasets.
 
\subsection{Joint-training Results on Detection and Segmentation}
\setlength{\tabcolsep}{4pt}
\begin{table}[t]
\caption{Joint-Training Performance of \methodNAME. Note that \textcolor{darkgreen}{Oracle} represents the dataset-specific training paradigm. We directly evaluate the generalist models on PACO to assess their recognition capability at the part level, as indicated by the results annotated in the \textcolor{lightgray}{grey} font.}
\begin{center}
\resizebox{0.99\columnwidth}{!}{
\begin{tabular}{llccccccccccccc}
\hline\noalign{\smallskip}
\multirow{3}{*}{Type} & \multirow{3}{*}{Method} & 
\multicolumn{7}{c}{Part-level Tasks} & \multicolumn{6}{c}{Object-level Tasks} \\
\cmidrule(lr){3-9} \cmidrule(lr){10-15}
& & \multicolumn{2}{c}{PartImageNet} & \multicolumn{2}{c}{Pascal Part} &\multicolumn{3}{c}{PACO} & \multicolumn{2}{c}{COCO-val} & \multicolumn{2}{c}{LVIS-minival} & \multicolumn{2}{c}{LVIS-val}  \\
\cmidrule(lr){3-4} \cmidrule(lr){5-6} \cmidrule(lr){7-9} \cmidrule(lr){10-11} \cmidrule(lr){12-13} \cmidrule(lr){14-15}
& & $\rm AP_{box}$ & $\rm AP_{mask} $& $\rm AP_{box}$ & $\rm AP_{mask} $ & $\rm AP_{mask}$ & $\rm AP_{mask}^{obj}$ & $\rm AP_{mask}^{opart}$ & $\rm AP_{box}$ & $\rm AP_{mask}$ & $\rm AP_{box}$ & $\rm AP_{mask}$ & $\rm AP_{box}$ & $\rm AP_{mask}$ \\
\noalign{\smallskip}
\hline
\noalign{\smallskip}
\multirow{9}{*}{Specialist}
& Mask2Former(R50)\cite{mask2former} & - & - & - & - & - & - & - & 46.2 & 43.7 & - & - & - & - \\
& Mask2Former(L)\cite{mask2former} & - & - & - & - & - & - & - & - & 50.1 & - & - & - & -  \\
& MaskDINO(R50)\cite{maskdino} & - & - & - & - & - & - & - & 50.5 & 46.0 & - & - & - & - \\
& MaskDINO(L)\cite{maskdino} & - & - & - & - & - & - & - & 58.3 & 52.1 & - & - & - & - \\
& ViTDet-L\cite{ViTDet} & - & - & - & - & - & - & - & 57.6 & 49.8 & - & - & 51.2 & 46.0 \\
& ViTDet-H\cite{ViTDet} & - & - & - & - & - & - & - & 57.6 & 49.8 & - & - & 53.4 & 48.1 \\
& EVA-02-L\cite{eva02} & - & - & - & - & - & - & - & 64.2 & 55.0 & - & - & 65.2 & 57.3\\
& PACO(R50)\cite{ramanathan2023paco} & - & - & - & - & - & 32.6 & 12.5 & - & - & - & - & - & -\\
& PACO(L)\cite{ramanathan2023paco} & - & - & - & - & - & 43.4 & 17.7 & - & - & - & - & - & -\\
\hline
\multirow{9}{*}{Generalist}
& Pix2Seq v2\cite{Pix2Seqv2} & - & - & - & - & - & - & - & 46.5 & 38.2 & - & - & - & - \\
& X-Decoder(L)\cite{xdecoder} & - & - & - & - 
& \textcolor{lightgray}{2.69} 
& \textcolor{lightgray}{11.9} 
& \textcolor{lightgray}{0.94} & - & 46.7 & - & - & - &- \\
& SEEM(L)\cite{SEEM} & - & - & - & - 
& \textcolor{lightgray}{1.99} 
& \textcolor{lightgray}{8.42} 
& \textcolor{lightgray}{0.69} & - & 47.7 & - & -  & - & - \\
& HIPIE(R50)\cite{wang2024hierarchical} & - & - & - & - & - & - & - & 53.9 & 45.9 & - & - & - & - \\
& Florence-2(B)\cite{florence2} & - & - & - & - & - & - & - & 41.4 & - & - & -  & - & -\\
& Florence-2(L)\cite{florence2} & - & - & - & - & - & - & - & 43.4 & - & -  & - & - & -  \\
& UNINEXT(R50)\cite{UNINEXT} & - & - & - & - & - & - & - &51.3 & 44.9 & - & - & 36.4 & -  \\
& UNINEXT(L)\cite{UNINEXT} & - & - & - & - & - & - & - & 58.1 &  49.6 & - & - & - & -\\
& GLEE(R50)\cite{wu2023GLEE} & - & - & - & - 
& \textcolor{lightgray}{3.44} 
& \textcolor{lightgray}{15.3} 
& \textcolor{lightgray}{1.29} & 55.0 & 48.4 & 50.5 & 45.9 & 44.2 & 40.2 \\
\hline
\multirow{6}{*}{Hierarchical}
 & VLPart(R50)\cite{peize2023vlpart} & 30.7 & 31.6 & 23.9 & 24.0 & 13.8 & 36.9 & 9.6 & 28.5 & - & - & 26.2 & - & -\\
 & \textcolor{darkgreen}{VLPart(R50)-Oracle\cite{peize2023vlpart}} & \textcolor{darkgreen}{29.2} & \textcolor{darkgreen}{29.7} & \textcolor{darkgreen}{18.9} & \textcolor{darkgreen}{19.4} & \textcolor{darkgreen}{13.3} & \textcolor{darkgreen}{28.0} & \textcolor{darkgreen}{10.6} & \textcolor{darkgreen}{38.0} & - & - & \textcolor{darkgreen}{28.1} & - & - \\
& VLPart(B)\cite{peize2023vlpart} & 43.9 & 41.2 & 33.5 & 31.7 & 22.1 & 55.0 & 15.9 & 40.3 & - & - & 39.6 & - & - \\
 & \textcolor{darkgreen}{VLPart(B)-Oracle\cite{peize2023vlpart}} & \textcolor{darkgreen}{44.3} & \textcolor{darkgreen}{41.7} & \textcolor{darkgreen}{29.2} & \textcolor{darkgreen}{27.4} & \textcolor{darkgreen}{19.1} & \textcolor{darkgreen}{37.7} & \textcolor{darkgreen}{15.2} & \textcolor{darkgreen}{52.5} & - & - & \textcolor{darkgreen}{43.1} & - & - \\
 \cline{2-15}
 & \cellcolor{gray!20}\methodNAME(R50) 
 & \cellcolor{gray!20}40.9
 & \cellcolor{gray!20}40.2 
 & \cellcolor{gray!20}35.0 
 & \cellcolor{gray!20}35.5
 & \cellcolor{gray!20}21.8
 & \cellcolor{gray!20}50.5
 & \cellcolor{gray!20}15.4
 & \cellcolor{gray!20}54.4 
 & \cellcolor{gray!20}47.6
 & \cellcolor{gray!20}48.7
 & \cellcolor{gray!20}43.5 
 & \cellcolor{gray!20}42.7
 & \cellcolor{gray!20}38.3 
 \\
 & \cellcolor{gray!20}\methodNAME(L) 
 & \cellcolor{gray!20}52.7 
 & \cellcolor{gray!20}50.9
 & \cellcolor{gray!20}39.6 
 & \cellcolor{gray!20}39.1 
 & \cellcolor{gray!20}27.8 
 & \cellcolor{gray!20}55.7 
 & \cellcolor{gray!20}21.3 
 & \cellcolor{gray!20}59.5 
 & \cellcolor{gray!20}52.0
 & \cellcolor{gray!20}56.5
 & \cellcolor{gray!20}50.6 
 & \cellcolor{gray!20}50.2 
 & \cellcolor{gray!20}45.0
 \\
\hline
\end{tabular}}
\end{center}
\label{table:joint_training_results}
\end{table}
\setlength{\tabcolsep}{1.4pt}
To endow our model with robust generalization capability, we perform joint training on various datasets and evaluate its performance on both object-level and part-level tasks. We compare our model with specialist and generalist models to evaluate its performance on object-level data. Additionally, we contrast it with VLPart to assess its performance on part-level datasets as well as the effectiveness of joint-training process on both types of datasets. As shown in \cref{table:joint_training_results}, \methodNAME significantly outperforms VLPart on both object-level and part-level tasks after joint-training, while achieving comparable performance on object-level tasks compared with previous SOTA. Through joint-training, our model has acquired strong generalization performance, allowing it to simultaneously address tasks for different hierarchies. We also observe that VLPart fails to achieve satisfactory performance on both object-level and part-level tasks. For example, VLPart obtains better performance on Pascal Part than its dataset-specific oracle, while decreasing its performance on COCO and LVIS. We attribute the performance drop of VLPart to the absence of hierarchical relationships, which causes confusion in modeling parts and objects and impairs object-level performance. \methodNAME effectively addresses this problem and extends the generalization capabilities from object-level to part-level tasks.

\subsection{Segmentation Result on SeginW Benchmark}
\setlength{\tabcolsep}{4pt}
\begin{table}[t]
\caption{Results on SeginW benchmark across 25 datasets. We report mAP.}
\begin{center}
\vspace{-2em}
\resizebox{1.0\columnwidth}{!}{
\begin{tabular}{l|l|ccccccccccccccccccccccccc}
Method & Mean & \rotatebox{90}{Airplane-Parts} & \rotatebox{90}{Bottles} &  \rotatebox{90}{Brain Tumor} & \rotatebox{90}{Chicken} & \rotatebox{90}{Cows} & \rotatebox{90}{Electric Shaver} & \rotatebox{90}{Elephants} & \rotatebox{90}{Fruits} & \rotatebox{90}{Garbage} & \rotatebox{90}{Ginger Garlic} & \rotatebox{90}{Hand} & \rotatebox{90}{Hand Metal} & \rotatebox{90}{House Parts} & \rotatebox{90}{HouseHold Items} & \rotatebox{90}{NutterflySquireel} & \rotatebox{90}{Phones} & \rotatebox{90}{Poles} & \rotatebox{90}{Puppies} & \rotatebox{90}{Rail} &  \rotatebox{90}{Salmon Fillet} & \rotatebox{90}{Strawberry} & \rotatebox{90}{Tablets} & \rotatebox{90}{Toolkits} & \rotatebox{90}{Trash} & \rotatebox{90}{Watermelon} \\
\hline
X-Decoder(L)\cite{xdecoder} & 32.3 & 13.1 & 42.1 & 2.2 & 8.6 & 44.9 & 7.5 & 66.0 &79.2 & 33.0 & 11.6 & 75.9 & 42.1 & 7.0 & 53.0 & 68.4 & 15.6 & 20.1 & 59.0 & 2.3 & 19.0 & 67.1 & 22.5 & 9.9 & 22.3 & 13.8 \\
OpenSEED(L)\cite{openseed} & 36.1 & 13.0 & 39.7 & 2.1 & 82.9 & 40.9 & 4.7 & 72.9 & 76.4 & 16.9 & 13.6 & 92.7 & 38.7 & 1.8 & 50.0 & 40.0 & 7.6 & 4.6 & 74.6 & 1.8 & 15.6 & 82.8 & 47.4 & 15.4 & 15.3 & 52.3 \\
ODISE(L)\cite{ODISE} & 38.7 & 15.8 & 37.7 & 2.9 & 84.1 & 41.6 & 18.3 & 74.9 & 81.3 & 39.8 & 23.0 & 41.4 & 51.4 & 9.3 & 60.4 & 71.9 & 43.8 & 0.4 & 65.4 & 2.8 & 30.2 & 79.9 & 9.1 & 15.0 & 28.6 & 37.5 \\
SAN(L)\cite{san} & 41.4 & 13.2 & 48.8 & 2.6 & 69.2 & 44.0 & 11.4 & 67.4 & 77.4 & 46.5 & 23.3 & 88.8 & 62.9 & 9.0 & 60.1 & 82.2 & 10.4 & 1.8 & 60.1 & 2.9 & 20.0 & 81.8 & 35.1 & 31.2 & 41.4 & 43.5\\
HIPIE(H)\cite{wang2024hierarchical} & 41.2 & 14.0 & 45.1 & 1.9 & 46.5 & 50.1 & 76.1 & 68.6 & 61.1 & 31.2 & 24.3 & 94.2 & 64.0 & 6.8 & 53.4 & 79.7 & 7.0 & 6.7 & 64.6 & 2.2 & 41.8 & 81.5 & 8.8 & 17.9 & 31.2 & 50.6 \\
UNINEXT(L)\cite{UNINEXT} & 42.1 & 15.1 & 46.1 & 2.6 & 75.2& 52.1 & 71.2 & 72.1 & 81.1 & 16.9 & 23.7 & 93.7 & 57.0 & 0.0 & 54.0 & 84.1 & 6.1 & 13.4 & 64.6 & 0.0 & 44.4 & 80.7 & 21.0 & 10.1 & 10.8 & 56.3\\
\rowcolor{gray!20} \methodNAME(R50) & 44.1 & 32.7 & 54.1 & 7.1 & 79.4 & 38.1 & 6.9 & 74.7 & 81.1 & 27.2 & 25.7 & 87.6 & 66.5 & 4.4 & 60.1 & 71.2 & 47.4 & 25.7 & 67.4 & 4.7 & 32.3 & 80.3 & 32.8 & 10.9 & 22.2 & 62.3 \\
\rowcolor{gray!20}\methodNAME(L) & 44.2 & 43.8 & 54.5 & 20.7 & 77.7 & 48.0 & 18.6 & 77.3 & 82.4 & 31.6 & 23.7 & 82.0 & 55.3 & 4.6 & 52.0 & 84.9 & 17.3 & 23.3 & 63.9 & 20.0 & 37.4 & 80.6 & 6.6 & 6.7 & 24.7 & 68.2\\
\hline
\end{tabular}}
\end{center}
\vspace{-2em}
\label{table:seginw_results}
\end{table}
\setlength{\tabcolsep}{1.4pt}

To further examine the zero-shot transferability of our model, we evaluate it on the Segmentation in the Wild (SeginW)\cite{xdecoder} which consists of 25 diverse segmentation datasets. Notably, during the inference process, we consider the House-Parts, Airplane-Parts, and Bottles datasets in SeginW as collections with part-level concepts. Hence, we conduct evaluation using hierarchical predictions for objects and parts. As for other datasets, we only utilize object-level predictions for testing. The results are shown in \cref{table:seginw_results}.

\subsection{Ablation Study}
\setlength{\tabcolsep}{4pt}
\begin{table}[t]
\caption{An ablation study on different model designs, as depicted in Fig~\ref{fig:ablation}. Note that Parallel Pixel Decoders refers to the utilization of two pixel decoders to generate feature maps at different hierarchies respectively. Independent Decoders denote the usage of two decoders, which facilitate the interaction between feature maps and queries at different hierarchies. Our final choice is scheme (c), which is highlighted in \sethlcolor{gray!20}\hl{gray}.}
\begin{center}
\vspace{-1.5em}
\resizebox{0.99\columnwidth}{!}{
\begin{tabular}{cccccccccccc}
\hline\noalign{\smallskip}
\multirow{3}{*}{Scheme} & \multicolumn{2}{c}{Model Design} & 
\multicolumn{5}{c}{Part-level Tasks} & \multicolumn{4}{c}{Object-level Tasks} \\

\cmidrule(lr){2-3} \cmidrule(lr){4-8} \cmidrule(lr){9-12}

& \multirow{2}{*}{\thead{Parrallel\\Pixel Decoders}} & \multirow{2}{*}{\thead{Independent\\Decoders}} & PartImageNet & Pascal Part &\multicolumn{3}{c}{PACO} & \multicolumn{2}{c}{COCO-val} & \multicolumn{2}{c}{LVIS-minival}  \\

\cmidrule(lr){4-4} \cmidrule(lr){5-5} \cmidrule(lr){6-8} \cmidrule(lr){9-10} \cmidrule(lr){11-12}

&  &  & $\rm AP_{mask} $ & $\rm AP_{mask} $ & $\rm AP_{mask}$ & $\rm AP_{mask}^{obj}$ & $\rm AP_{mask}^{opart}$ & $\rm AP_{box}$ & $\rm AP_{mask}$ & $\rm AP_{box}$ & $\rm AP_{mask}$ \\

\noalign{\smallskip}
\hline
\noalign{\smallskip}
(a) & \usym{2613} & \usym{2613} & \textbf{39.0} & 34.1 & 20.1 & 47.4 & 13.5 & 47.8 & 43.5 & 34.8 & 33.4  \\
(b) & \checkmark & \usym{2613}  & 38.3 & 34.5 & 20.8 & \textbf{48.8} & 13.8 & 48.5 & \textbf{44.3} & 34.9 & \textbf{34.2}  \\
\rowcolor{gray!20} (c) & \usym{2613} & \checkmark  & \textbf{39.0} & \textbf{34.7} & \textbf{20.9} & 47.9 & \textbf{14.2} & \textbf{49.3} & 44.2 & \textbf{35.6} & 33.8 \\
\hline
\end{tabular}}
\end{center}
\label{table:ablation_design}
\vspace{-1em}
\end{table}
\setlength{\tabcolsep}{1.4pt}
\begin{figure*}[tb]
\centering
\includegraphics[width=0.99 \linewidth]{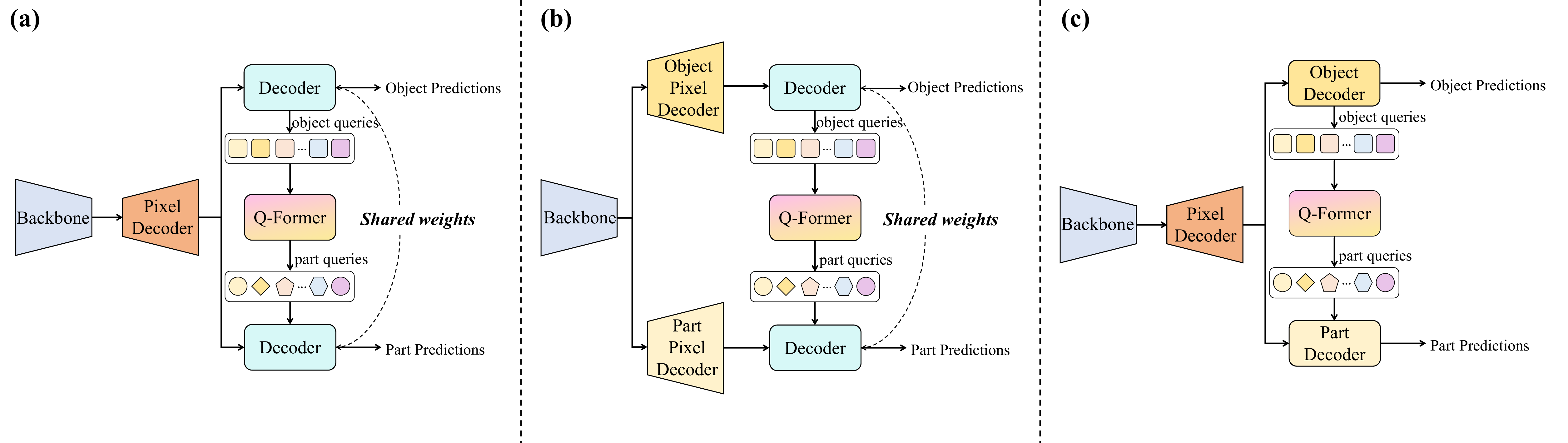}
\vspace{-2ex}
\caption{
Various designs for generating predictions at different hierarchies. In scheme (a), we only utilize a single decoder to generate predictions for both objects and parts. In scheme (b), two parallel pixel decoders are employed to generate feature maps at different levels, aiming to explore the effectiveness of feature maps at different granularity. In scheme (c), we use two independent decoders to generate predictions for objects and parts respectively.
}
\label{fig:ablation}
\vspace{-2ex}
\end{figure*}

To demonstrate that our model design achieves satisfactory results on both object-level and part-level tasks, we conduct an ablation study (depicted in \cref{fig:ablation}) on the model architecture and present results in \cref{table:ablation_design}. We ablate with a backbone of ResNet-50 and perform joint-training on COCO\cite{coco}, LVIS\cite{lvis}, PartImageNet\cite{he2021partimagenet}, Pascal Part\cite{chen2014detect} and PACO\cite{ramanathan2023paco} with 90K iterations. From this study, we draw several important conclusions: (1) The utilization of parallel pixel decoders only results in slight improvements in mask predictions on few datasets, indicating that the influence of feature maps at different granularities is negligible. (2) Adopting independent decoders to obtain predictions at different levels demonstrates superior performance across the majority of datasets, manifesting the effectiveness of independent decoders.
As adopting parallel pixel decoders (b) results in significant GPU memory costs without considerable gains, and all metrics for (a) are lower than (c), we select (c) as our final model design.
Additional ablation studies, extensive qualitative analysis, and experiments on mLLM can be found in the appendix.

\subsection{Limitations}
In this work, we still adopt CLIP as the text encoder, which is trained on text-image pairs and thus lacks the ability to perceive fine-grained descriptions of object or part instances. This limitation may restrict the improvement of model performance and prompts us to consider how to enhance the perception capabilities of region-level models, which will be our future work.


\section{Conclusion}
In this paper, we introduce \methodNAME, a groundbreaking foundation model designed towards a complete comprehension of both objects and parts in images. 
Through the generic hierarchical relationships established by 
the Q-Former, we are able to break through the limitation of scarce part-level data by introducing a large amount of object-level data, thereby transferring the powerful generalization capabilities from objects to parts.
Through extensive training on diverse datasets, \methodNAME achieves SOTA performance across various part-level tasks while maintaining competitive results on object-level tasks, enabling it to parse any objects into parts and serve as a foundation model for general fine-grained region-level perception tasks.

\section*{Acknowledgements}
This work was supported by the National Science Fund for Distinguished Young Scholars of China (Grant No.62225603).

\clearpage  

%
%
\bibliographystyle{splncs04}
\bibliography{egbib}

\begin{thebibliography}{100}
\providecommand{\url}[1]{\texttt{#1}}
\providecommand{\urlprefix}{URL }
\providecommand{\doi}[1]{https://doi.org/#1}

\bibitem{flamingo}
Alayrac, J.B., Donahue, J., Luc, P., Miech, A., Barr, I., Hasson, Y., Lenc, K., Mensch, A., Millican, K., Reynolds, M., et~al.: Flamingo: a visual language model for few-shot learning. In: Advances in neural information processing systems. vol.~35, pp. 23716--23736 (2022)

\bibitem{brohan2022rt}
Brohan, A., Brown, N., Carbajal, J., Chebotar, Y., Dabis, J., Finn, C., Gopalakrishnan, K., Hausman, K., Herzog, A., Hsu, J., et~al.: Rt-1: Robotics transformer for real-world control at scale. arXiv preprint arXiv:2212.06817  (2022)

\bibitem{gpt3}
Brown, T., Mann, B., Ryder, N., Subbiah, M., Kaplan, J.D., Dhariwal, P., Neelakantan, A., Shyam, P., Sastry, G., Askell, A., et~al.: Language models are few-shot learners. In: Advances in neural information processing systems. vol.~33, pp. 1877--1901 (2020)

\bibitem{sslDINO}
Caron, M., Touvron, H., Misra, I., J{\'e}gou, H., Mairal, J., Bojanowski, P., Joulin, A.: Emerging properties in self-supervised vision transformers. In: Proceedings of the IEEE/CVF international conference on computer vision. pp. 9650--9660 (2021)

\bibitem{Pix2Seqv2}
Chen, T., Saxena, S., Li, L., Lin, T.Y., Fleet, D.J., Hinton, G.E.: A unified sequence interface for vision tasks. Advances in Neural Information Processing Systems  \textbf{35},  31333--31346 (2022)

\bibitem{chen2014detect}
Chen, X., Mottaghi, R., Liu, X., Fidler, S., Urtasun, R., Yuille, A.: Detect what you can: Detecting and representing objects using holistic models and body parts. In: Proceedings of the IEEE conference on computer vision and pattern recognition. pp. 1971--1978 (2014)

\bibitem{mask2former}
Cheng, B., Misra, I., Schwing, A.G., Kirillov, A., Girdhar, R.: Masked-attention mask transformer for universal image segmentation. In: Proceedings of the IEEE/CVF conference on computer vision and pattern recognition. pp. 1290--1299 (2022)

\bibitem{imagenet}
Deng, J., Dong, W., Socher, R., Li, L.J., Li, K., Fei-Fei, L.: Imagenet: A large-scale hierarchical image database. In: Proceedings of the IEEE/CVF International Conference on Computer Vision. pp. 248--255 (2009)

\bibitem{bert}
Devlin, J., Chang, M.W., Lee, K., Toutanova, K.: Bert: Pre-training of deep bidirectional transformers for language understanding. arXiv preprint arXiv:1810.04805  (2018)

\bibitem{Dong_2014_CVPR}
Dong, J., Chen, Q., Shen, X., Yang, J., Yan, S.: Towards unified human parsing and pose estimation. In: Proceedings of the IEEE Conference on Computer Vision and Pattern Recognition. pp. 843--850 (2014)

\bibitem{vqgan}
Esser, P., Rombach, R., Ommer, B.: Taming transformers for high-resolution image synthesis. In: Proceedings of the IEEE/CVF conference on computer vision and pattern recognition. pp. 12873--12883 (2021)

\bibitem{eva02}
Fang, Y., Sun, Q., Wang, X., Huang, T., Wang, X., Cao, Y.: Eva-02: A visual representation for neon genesis. arXiv preprint arXiv:2303.11331  (2023)

\bibitem{fang2023eva}
Fang, Y., Wang, W., Xie, B., Sun, Q., Wu, L., Wang, X., Huang, T., Wang, X., Cao, Y.: Eva: Exploring the limits of masked visual representation learning at scale. In: Proceedings of the IEEE/CVF Conference on Computer Vision and Pattern Recognition. pp. 19358--19369 (2023)

\bibitem{degeus2021panopticparts}
de~Geus, D., Meletis, P., Lu, C., Wen, X., Dubbelman, G.: Part-aware panoptic segmentation. In: Proceedings of the IEEE/CVF Conference on Computer Vision and Pattern Recognition. pp. 5485--5494 (2021)

\bibitem{gong2017look}
Gong, K., Liang, X., Zhang, D., Shen, X., Lin, L.: Look into person: Self-supervised structure-sensitive learning and a new benchmark for human parsing. In: Proceedings of the IEEE conference on computer vision and pattern recognition. pp. 932--940 (2017)

\bibitem{lvis}
Gupta, A., Dollar, P., Girshick, R.: Lvis: A dataset for large vocabulary instance segmentation. In: Proceedings of the IEEE/CVF conference on computer vision and pattern recognition. pp. 5356--5364 (2019)

\bibitem{He_2023_CVPR}
He, J., Chen, J., Lin, M.X., Yu, Q., Yuille, A.L.: Compositor: Bottom-up clustering and compositing for robust part and object segmentation. In: Proceedings of the IEEE/CVF Conference on Computer Vision and Pattern Recognition. pp. 11259--11268 (2023)

\bibitem{he2021partimagenet}
He, J., Yang, S., Yang, S., Kortylewski, A., Yuan, X., Chen, J.N., Liu, S., Yang, C., Yu, Q., Yuille, A.: Partimagenet: A large, high-quality dataset of parts. In: European Conference on Computer Vision. pp. 128--145. Springer (2022)

\bibitem{MAE}
He, K., Chen, X., Xie, S., Li, Y., Doll{\'a}r, P., Girshick, R.: Masked autoencoders are scalable vision learners. In: Proceedings of the IEEE/CVF conference on computer vision and pattern recognition. pp. 16000--16009 (2022)

\bibitem{resnet}
He, K., Zhang, X., Ren, S., Sun, J.: Deep residual learning for image recognition. In: Proceedings of the IEEE conference on computer vision and pattern recognition. pp. 770--778 (2016)

\bibitem{ALIGN}
Jia, C., Yang, Y., Xia, Y., Chen, Y.T., Parekh, Z., Pham, H., Le, Q., Sung, Y.H., Li, Z., Duerig, T.: Scaling up visual and vision-language representation learning with noisy text supervision. In: International conference on machine learning. pp. 4904--4916. PMLR (2021)

\bibitem{mdetr}
Kamath, A., Singh, M., LeCun, Y., Synnaeve, G., Misra, I., Carion, N.: Mdetr-modulated detection for end-to-end multi-modal understanding. In: ICCV. pp. 1780--1790 (2021)

\bibitem{kawar2023imagic}
Kawar, B., Zada, S., Lang, O., Tov, O., Chang, H., Dekel, T., Mosseri, I., Irani, M.: Imagic: Text-based real image editing with diffusion models. In: Proceedings of the IEEE/CVF Conference on Computer Vision and Pattern Recognition. pp. 6007--6017 (2023)

\bibitem{SAM}
Kirillov, A., Mintun, E., Ravi, N., Mao, H., Rolland, C., Gustafson, L., Xiao, T., Whitehead, S., Berg, A.C., Lo, W.Y., et~al.: Segment anything. In: Proceedings of the IEEE/CVF International Conference on Computer Vision. pp. 4015--4026 (2023)

\bibitem{visualgenome}
Krishna, R., Zhu, Y., Groth, O., Johnson, J., Hata, K., Kravitz, J., Chen, S., Kalantidis, Y., Li, L.J., Shamma, D.A., et~al.: Visual genome: Connecting language and vision using crowdsourced dense image annotations. International journal of computer vision  \textbf{123},  32--73 (2017)

\bibitem{OpenImages}
Kuznetsova, A., Rom, H., Alldrin, N., Uijlings, J., Krasin, I., Pont-Tuset, J., Kamali, S., Popov, S., Malloci, M., Kolesnikov, A., et~al.: The open images dataset v4: Unified image classification, object detection, and visual relationship detection at scale. International journal of computer vision  \textbf{128}(7),  1956--1981 (2020)

\bibitem{li2023semantic}
Li, F., Zhang, H., Sun, P., Zou, X., Liu, S., Yang, J., Li, C., Zhang, L., Gao, J.: Semantic-sam: Segment and recognize anything at any granularity. arXiv preprint arXiv:2307.04767  (2023)

\bibitem{maskdino}
Li, F., Zhang, H., Xu, H., Liu, S., Zhang, L., Ni, L.M., Shum, H.Y.: Mask dino: Towards a unified transformer-based framework for object detection and segmentation. In: Proceedings of the IEEE/CVF Conference on Computer Vision and Pattern Recognition. pp. 3041--3050 (2023)

\bibitem{Uni-perceiverv2}
Li, H., Zhu, J., Jiang, X., Zhu, X., Li, H., Yuan, C., Wang, X., Qiao, Y., Wang, X., Wang, W., et~al.: Uni-perceiver v2: A generalist model for large-scale vision and vision-language tasks. In: Proceedings of the IEEE/CVF Conference on Computer Vision and Pattern Recognition. pp. 2691--2700 (2023)

\bibitem{li2017multiple}
Li, J., Zhao, J., Wei, Y., Lang, C., Li, Y., Sim, T., Yan, S., Feng, J.: Multiple-human parsing in the wild. arXiv preprint arXiv:1705.07206  (2017)

\bibitem{GLIP}
Li, L.H., Zhang, P., Zhang, H., Yang, J., Li, C., Zhong, Y., Wang, L., Yuan, L., Zhang, L., Hwang, J.N., et~al.: Grounded language-image pre-training. In: Proceedings of the IEEE/CVF Conference on Computer Vision and Pattern Recognition. pp. 10965--10975 (2022)

\bibitem{li2022panoptic}
Li, X., Xu, S., Yang, Y., Cheng, G., Tong, Y., Tao, D.: Panoptic-partformer: Learning a unified model for panoptic part segmentation. In: European Conference on Computer Vision. pp. 729--747. Springer (2022)

\bibitem{ViTDet}
Li, Y., Mao, H., Girshick, R., He, K.: Exploring plain vision transformer backbones for object detection. In: European Conference on Computer Vision. pp. 280--296. Springer (2022)

\bibitem{li2021partgan}
Li, Y., Singh, K.K., Xue, Y., Lee, Y.J.: Partgan: Weakly-supervised part decomposition for image generation and segmentation. In: British Machine Vision Conference (BMVC) (2021)

\bibitem{lin2024generative}
Lin, C., Jiang, Y., Qu, L., Yuan, Z., Cai, J.: Generative region-language pretraining for open-ended object detection. In: Proceedings of the IEEE/CVF Conference on Computer Vision and Pattern Recognition. pp. 13958--13968 (2024)

\bibitem{lin2022learning}
Lin, C., Sun, P., Jiang, Y., Luo, P., Qu, L., Haffari, G., Yuan, Z., Cai, J.: Learning object-language alignments for open-vocabulary object detection. In: The Eleventh International Conference on Learning Representations (2023)

\bibitem{focalloss}
Lin, T.Y., Goyal, P., Girshick, R., He, K., Doll{\'a}r, P.: Focal loss for dense object detection. In: Proceedings of the IEEE international conference on computer vision. pp. 2980--2988 (2017)

\bibitem{coco}
Lin, T.Y., Maire, M., Belongie, S., Hays, J., Perona, P., Ramanan, D., Doll{\'a}r, P., Zitnick, C.L.: Microsoft coco: Common objects in context. In: European Conference on Computer Vision. pp. 740--755. Springer (2014)

\bibitem{ling2021editgan}
Ling, H., Kreis, K., Li, D., Kim, S.W., Torralba, A., Fidler, S.: Editgan: High-precision semantic image editing. In: Advances in Neural Information Processing Systems. vol.~34, pp. 16331--16345 (2021)

\bibitem{polyformer}
Liu, J., Ding, H., Cai, Z., Zhang, Y., Satzoda, R.K., Mahadevan, V., Manmatha, R.: Polyformer: Referring image segmentation as sequential polygon generation. In: Proceedings of the IEEE/CVF Conference on Computer Vision and Pattern Recognition. pp. 18653--18663 (2023)

\bibitem{liu2023grounding}
Liu, S., Zeng, Z., Ren, T., Li, F., Zhang, H., Yang, J., Li, C., Yang, J., Su, H., Zhu, J., et~al.: Grounding dino: Marrying dino with grounded pre-training for open-set object detection. arXiv preprint arXiv:2303.05499  (2023)

\bibitem{SwinTransformer}
Liu, Z., Lin, Y., Cao, Y., Hu, H., Wei, Y., Zhang, Z., Lin, S., Guo, B.: Swin transformer: Hierarchical vision transformer using shifted windows. In: Proceedings of the IEEE/CVF international conference on computer vision. pp. 10012--10022 (2021)

\bibitem{AdamW}
Loshchilov, I., Hutter, F.: Decoupled weight decay regularization. In: International Conference on Learning Representations (2019)

\bibitem{unified-io}
Lu, J., Clark, C., Zellers, R., Mottaghi, R., Kembhavi, A.: Unified-io: A unified model for vision, language, and multi-modal tasks. In: The Eleventh International Conference on Learning Representations (2022)

\bibitem{codet}
Ma, C., Jiang, Y., Wen, X., Yuan, Z., Qi, X.: Codet: Co-occurrence guided region-word alignment for open-vocabulary object detection. In: Advances in neural information processing systems. vol.~36 (2023)

\bibitem{ma2024groma}
Ma, C., Jiang, Y., Wu, J., Yuan, Z., Qi, X.: Groma: Localized visual tokenization for grounding multimodal large language models. arXiv preprint arXiv:2404.13013  (2024)

\bibitem{RefCOCOg-umd}
Mao, J., Huang, J., Toshev, A., Camburu, O., Yuille, A.L., Murphy, K.: Generation and comprehension of unambiguous object descriptions. In: Proceedings of the IEEE conference on computer vision and pattern recognition. pp. 11--20 (2016)

\bibitem{meletis2020cityscapes}
Meletis, P., Wen, X., Lu, C., de~Geus, D., Dubbelman, G.: Cityscapes-panoptic-parts and pascal-panoptic-parts datasets for scene understanding. arXiv preprint arXiv:2004.07944  (2020)

\bibitem{michieli2020gmnet}
Michieli, U., Borsato, E., Rossi, L., Zanuttigh, P.: Gmnet: Graph matching network for large scale part semantic segmentation in the wild. In: European Conference on Computer Vision. pp. 397--414. Springer (2020)

\bibitem{diceloss}
Milletari, F., Navab, N., Ahmadi, S.A.: V-net: Fully convolutional neural networks for volumetric medical image segmentation. In: 2016 fourth international conference on 3D vision (3DV) (2016)

\bibitem{morabia2020attention}
Morabia, K., Arora, J., Vijaykumar, T.: Attention-based joint detection of object and semantic part. arXiv preprint arXiv:2007.02419  (2020)

\bibitem{nair2022r3m}
Nair, S., Rajeswaran, A., Kumar, V., Finn, C., Gupta, A.: R3m: A universal visual representation for robot manipulation. In: Conference on Robot Learning. pp. 892--909. PMLR (2023)

\bibitem{ng2022animal}
Ng, X.L., Ong, K.E., Zheng, Q., Ni, Y., Yeo, S.Y., Liu, J.: Animal kingdom: A large and diverse dataset for animal behavior understanding. In: Proceedings of the IEEE/CVF Conference on Computer Vision and Pattern Recognition. pp. 19023--19034 (2022)

\bibitem{pan2023towards}
Pan, T.Y., Liu, Q., Chao, W.L., Price, B.: Towards open-world segmentation of parts. In: Proceedings of the IEEE/CVF Conference on Computer Vision and Pattern Recognition. pp. 15392--15401 (2023)

\bibitem{OVIS}
Qi, J., Gao, Y., Hu, Y., Wang, X., Liu, X., Bai, X., Belongie, S., Yuille, A., Torr, P.H., Bai, S.: Occluded video instance segmentation: A benchmark. International Journal of Computer Vision  \textbf{130}(8),  2022--2039 (2022)

\bibitem{qi2024aims}
Qi, L., Kuen, J., Guo, W., Gu, J., Lin, Z., Du, B., Xu, Y., Yang, M.H.: Aims: All-inclusive multi-level segmentation for anything. In: Advances in Neural Information Processing Systems. vol.~36 (2023)

\bibitem{CLIP}
Radford, A., Kim, J.W., Hallacy, C., Ramesh, A., Goh, G., Agarwal, S., Sastry, G., Askell, A., Mishkin, P., Clark, J., et~al.: Learning transferable visual models from natural language supervision. In: International conference on machine learning. pp. 8748--8763. PMLR (2021)

\bibitem{T5}
Raffel, C., Shazeer, N., Roberts, A., Lee, K., Narang, S., Matena, M., Zhou, Y., Li, W., Liu, P.J.: Exploring the limits of transfer learning with a unified text-to-text transformer. The Journal of Machine Learning Research  \textbf{21}(1),  5485--5551 (2020)

\bibitem{ramanathan2023paco}
Ramanathan, V., Kalia, A., Petrovic, V., Wen, Y., Zheng, B., Guo, B., Wang, R., Marquez, A., Kovvuri, R., Kadian, A., et~al.: Paco: Parts and attributes of common objects. In: Proceedings of the IEEE/CVF Conference on Computer Vision and Pattern Recognition. pp. 7141--7151 (2023)

\bibitem{DALLE2}
Ramesh, A., Dhariwal, P., Nichol, A., Chu, C., Chen, M.: Hierarchical text-conditional image generation with clip latents. arXiv preprint arXiv:2204.06125  \textbf{1}(2), ~3 (2022)

\bibitem{DALLE}
Ramesh, A., Pavlov, M., Goh, G., Gray, S., Voss, C., Radford, A., Chen, M., Sutskever, I.: Zero-shot text-to-image generation. In: International Conference on Machine Learning. pp. 8821--8831. PMLR (2021)

\bibitem{reddy2018carfusion}
Reddy, N.D., Vo, M., Narasimhan, S.G.: Carfusion: Combining point tracking and part detection for dynamic 3d reconstruction of vehicles. In: Proceedings of the IEEE conference on computer vision and pattern recognition. pp. 1906--1915 (2018)

\bibitem{giou}
Rezatofighi, H., Tsoi, N., Gwak, J., Sadeghian, A., Reid, I., Savarese, S.: Generalized intersection over union: A metric and a loss for bounding box regression. In: Proceedings of the IEEE/CVF conference on computer vision and pattern recognition. pp. 658--666 (2019)

\bibitem{stable_diffusion}
Rombach, R., Blattmann, A., Lorenz, D., Esser, P., Ommer, B.: High-resolution image synthesis with latent diffusion models. In: Proceedings of the IEEE/CVF conference on computer vision and pattern recognition. pp. 10684--10695 (2022)

\bibitem{objects365}
Shao, S., Li, Z., Zhang, T., Peng, C., Yu, G., Zhang, X., Li, J., Sun, J.: Objects365: A large-scale, high-quality dataset for object detection. In: Proceedings of the IEEE/CVF international conference on computer vision. pp. 8430--8439 (2019)

\bibitem{peize2023vlpart}
Sun, P., Chen, S., Zhu, C., Xiao, F., Luo, P., Xie, S., Yan, Z.: Going denser with open-vocabulary part segmentation. In: Proceedings of the IEEE/CVF International Conference on Computer Vision. pp. 15453--15465 (2023)

\bibitem{tang2022request}
Tang, C., Xie, L., Zhang, X., Hu, X., Tian, Q.: Visual recognition by request. In: Proceedings of the IEEE/CVF Conference on Computer Vision and Pattern Recognition. pp. 15265--15274 (2023)

\bibitem{WahCUB_200_2011}
Wah, C., Branson, S., Welinder, P., Perona, P., Belongie, S.: Caltech-ucsd birds-200-2011 (cub-200-2011). Tech. Rep. CNS-TR-2011-001, California Institute of Technology (2011)

\bibitem{wang2015joint}
Wang, P., Shen, X., Lin, Z., Cohen, S., Price, B., Yuille, A.L.: Joint object and part segmentation using deep learned potentials. In: Proceedings of the IEEE International Conference on Computer Vision. pp. 1573--1581 (2015)

\bibitem{OFA}
Wang, P., Yang, A., Men, R., Lin, J., Bai, S., Li, Z., Ma, J., Zhou, C., Zhou, J., Yang, H.: Ofa: Unifying architectures, tasks, and modalities through a simple sequence-to-sequence learning framework. In: International Conference on Machine Learning. pp. 23318--23340. PMLR (2022)

\bibitem{UVO}
Wang, W., Feiszli, M., Wang, H., Tran, D.: Unidentified video objects: A benchmark for dense, open-world segmentation. In: Proceedings of the IEEE/CVF International Conference on Computer Vision. pp. 10776--10785 (2021)

\bibitem{beit3}
Wang, W., Bao, H., Dong, L., Bjorck, J., Peng, Z., Liu, Q., Aggarwal, K., Mohammed, O.K., Singhal, S., Som, S., et~al.: Image as a foreign language: Beit pretraining for all vision and vision-language tasks. arXiv preprint arXiv:2208.10442  (2022)

\bibitem{wang2024hierarchical}
Wang, X., Li, S., Kallidromitis, K., Kato, Y., Kozuka, K., Darrell, T.: Hierarchical open-vocabulary universal image segmentation. In: Advances in Neural Information Processing Systems. vol.~36 (2023)

\bibitem{deepflux}
Wang, Y., Xu, Y., Tsogkas, S., Bai, X., Dickinson, S., Siddiqi, K.: Deepflux for skeletons in the wild. In: Proceedings of the IEEE/CVF conference on computer vision and pattern recognition. pp. 5287--5296 (2019)

\bibitem{wei2023ov}
Wei, M., Yue, X., Zhang, W., Kong, S., Liu, X., Pang, J.: Ov-parts: Towards open-vocabulary part segmentation. In: Thirty-seventh Conference on Neural Information Processing Systems Datasets and Benchmarks Track (2023)

\bibitem{wu2023GLEE}
Wu, J., Jiang, Y., Liu, Q., Yuan, Z., Bai, X., Bai, S.: General object foundation model for images and videos at scale. In: Proceedings of the IEEE/CVF Conference on Computer Vision and Pattern Recognition. pp. 3783--3795 (2024)

\bibitem{florence2}
Xiao, B., Wu, H., Xu, W., Dai, X., Hu, H., Lu, Y., Zeng, M., Liu, C., Yuan, L.: Florence-2: Advancing a unified representation for a variety of vision tasks. In: Proceedings of the IEEE/CVF Conference on Computer Vision and Pattern Recognition. pp. 4818--4829 (2024)

\bibitem{ODISE}
Xu, J., Liu, S., Vahdat, A., Byeon, W., Wang, X., De~Mello, S.: Open-vocabulary panoptic segmentation with text-to-image diffusion models. In: Proceedings of the IEEE/CVF Conference on Computer Vision and Pattern Recognition. pp. 2955--2966 (2023)

\bibitem{san}
Xu, M., Zhang, Z., Wei, F., Hu, H., Bai, X.: Side adapter network for open-vocabulary semantic segmentation. In: Proceedings of the IEEE/CVF Conference on Computer Vision and Pattern Recognition (CVPR). pp. 2945--2954 (June 2023)

\bibitem{ytvis21dataset}
Xu, N., Yang, L., Yang, J., Yue, D., Fan, Y., Liang, Y., Huang., T.S.: Youtubevis dataset 2021 version. \url{https://youtube-vos.org/dataset/vis/}

\bibitem{unicorn}
Yan, B., Jiang, Y., Sun, P., Wang, D., Yuan, Z., Luo, P., Lu, H.: Towards grand unification of object tracking. In: European Conference on Computer Vision (2022)

\bibitem{UNINEXT}
Yan, B., Jiang, Y., Wu, J., Wang, D., Luo, P., Yuan, Z., Lu, H.: Universal instance perception as object discovery and retrieval. In: Proceedings of the IEEE/CVF Conference on Computer Vision and Pattern Recognition. pp. 15325--15336 (2023)

\bibitem{SoM}
Yang, J., Zhang, H., Li, F., Zou, X., Li, C., Gao, J.: Set-of-mark prompting unleashes extraordinary visual grounding in gpt-4v. arXiv preprint arXiv:2310.11441  (2023)

\bibitem{ytvis2019}
Yang, L., Fan, Y., Xu, N.: Video instance segmentation. In: Proceedings of the IEEE/CVF International Conference on Computer Vision. pp. 5188--5197 (2019)

\bibitem{yang2019parsing}
Yang, L., Song, Q., Wang, Z., Jiang, M.: Parsing r-cnn for instance-level human analysis. In: Proceedings of the IEEE/CVF conference on computer vision and pattern recognition. pp. 364--373 (2019)

\bibitem{yang2011articulated}
Yang, Y., Ramanan, D.: Articulated pose estimation with flexible mixtures-of-parts. In: Proceedings of the IEEE/CVF Conference on Computer Vision and Pattern Recognition. pp. 1385--1392. IEEE (2011)

\bibitem{unitab}
Yang, Z., Gan, Z., Wang, J., Hu, X., Ahmed, F., Liu, Z., Lu, Y., Wang, L.: Unitab: Unifying text and box outputs for grounded vision-language modeling. In: European Conference on Computer Vision. pp. 521--539. Springer (2022)

\bibitem{yao2023detclipv2}
Yao, L., Han, J., Liang, X., Xu, D., Zhang, W., Li, Z., Xu, H.: Detclipv2: Scalable open-vocabulary object detection pre-training via word-region alignment. In: Proceedings of the IEEE/CVF Conference on Computer Vision and Pattern Recognition. pp. 23497--23506 (2023)

\bibitem{detclip}
Yao, L., Han, J., Wen, Y., Liang, X., Xu, D., Zhang, W., Li, Z., Xu, C., Xu, H.: Detclip: Dictionary-enriched visual-concept paralleled pre-training for open-world detection. In: Advances in Neural Information Processing Systems. vol.~35, pp. 9125--9138 (2022)

\bibitem{bdd100k}
Yu, F., Chen, H., Wang, X., Xian, W., Chen, Y., Liu, F., Madhavan, V., Darrell, T.: Bdd100k: A diverse driving dataset for heterogeneous multitask learning. In: Proceedings of the IEEE/CVF conference on computer vision and pattern recognition. pp. 2636--2645 (2020)

\bibitem{RefCOCOandplus}
Yu, L., Poirson, P., Yang, S., Berg, A.C., Berg, T.L.: Modeling context in referring expressions. In: European Conference on Computer Vision. pp. 69--85. Springer (2016)

\bibitem{florence}
Yuan, L., Chen, D., Chen, Y.L., Codella, N., Dai, X., Gao, J., Hu, H., Huang, X., Li, B., Li, C., et~al.: Florence: A new foundation model for computer vision. arXiv preprint arXiv:2111.11432  (2021)

\bibitem{zareian2021open}
Zareian, A., Rosa, K.D., Hu, D.H., Chang, S.F.: Open-vocabulary object detection using captions. In: Proceedings of the IEEE/CVF Conference on Computer Vision and Pattern Recognition. pp. 14393--14402 (2021)

\bibitem{openseed}
Zhang, H., Li, F., Zou, X., Liu, S., Li, C., Yang, J., Zhang, L.: A simple framework for open-vocabulary segmentation and detection. In: Proceedings of the IEEE/CVF International Conference on Computer Vision. pp. 1020--1031 (2023)

\bibitem{GLIPv2}
Zhang, H., Zhang, P., Hu, X., Chen, Y.C., Li, L., Dai, X., Wang, L., Yuan, L., Hwang, J.N., Gao, J.: Glipv2: Unifying localization and vision-language understanding. In: Koyejo, S., Mohamed, S., Agarwal, A., Belgrave, D., Cho, K., Oh, A. (eds.) Advances in Neural Information Processing Systems. vol.~35, pp. 36067--36080 (2022)

\bibitem{zhong2022regionclip}
Zhong, Y., Yang, J., Zhang, P., Li, C., Codella, N., Li, L.H., Zhou, L., Dai, X., Yuan, L., Li, Y., et~al.: Regionclip: Region-based language-image pretraining. In: Proceedings of the IEEE/CVF Conference on Computer Vision and Pattern Recognition. pp. 16793--16803 (2022)

\bibitem{zhou2018semantic}
Zhou, B., Zhao, H., Puig, X., Xiao, T., Fidler, S., Barriuso, A., Torralba, A.: Semantic understanding of scenes through the ade20k dataset. International Journal on Computer Vision  (2018)

\bibitem{zhou2021differentiable}
Zhou, T., Wang, W., Liu, S., Yang, Y., Van~Gool, L.: Differentiable multi-granularity human representation learning for instance-aware human semantic parsing. In: Proceedings of the IEEE/CVF conference on computer vision and pattern recognition. pp. 1622--1631 (2021)

\bibitem{seqtr}
Zhu, C., Zhou, Y., Shen, Y., Luo, G., Pan, X., Lin, M., Chen, C., Cao, L., Sun, X., Ji, R.: Seqtr: A simple yet universal network for visual grounding. In: European Conference on Computer Vision. pp. 598--615. Springer (2022)

\bibitem{deformableDETR}
Zhu, X., Su, W., Lu, L., Li, B., Wang, X., Dai, J.: Deformable detr: Deformable transformers for end-to-end object detection. In: International Conference on Learning Representations (2021)

\bibitem{Uni-perceiver}
Zhu, X., Zhu, J., Li, H., Wu, X., Li, H., Wang, X., Dai, J.: Uni-perceiver: Pre-training unified architecture for generic perception for zero-shot and few-shot tasks. In: Proceedings of the IEEE/CVF Conference on Computer Vision and Pattern Recognition. pp. 16804--16815 (2022)

\bibitem{ziegler2022self}
Ziegler, A., Asano, Y.M.: Self-supervised learning of object parts for semantic segmentation. In: Proceedings of the IEEE/CVF Conference on Computer Vision and Pattern Recognition. pp. 14502--14511 (2022)

\bibitem{xdecoder}
Zou, X., Dou, Z.Y., Yang, J., Gan, Z., Li, L., Li, C., Dai, X., Behl, H., Wang, J., Yuan, L., et~al.: Generalized decoding for pixel, image, and language. In: Proceedings of the IEEE/CVF Conference on Computer Vision and Pattern Recognition. pp. 15116--15127 (2023)

\bibitem{SEEM}
Zou, X., Yang, J., Zhang, H., Li, F., Li, L., Wang, J., Wang, L., Gao, J., Lee, Y.J.: Segment everything everywhere all at once. In: Advances in Neural Information Processing Systems. vol.~36 (2023)

\end{thebibliography}

\clearpage  

\appendix
\renewcommand{\thetable}{\Roman{table}}
\renewcommand{\thefigure}{\Roman{figure}}

\title{PartGLEE: A Foundation Model for Recognizing and Parsing Any Objects \\ Appendix}
\titlerunning{PartGLEE}
\author{Junyi Li\inst{1*} \and
Junfeng Wu\inst{1*} \and
Weizhi Zhao\inst{1} \and
Song Bai\inst{2} \and 
Xiang Bai\inst{1\dag}
}
\authorrunning{J. Li, J. Wu, et al.}
\institute{
Huazhong University of Science and Technology \and ByteDance Inc. 
}

\maketitle

In the appendix, we first present additional evaluation results of our model in \cref{sec:supp_additional_results} and then provide more detailed information on data unification as well as training strategies in \cref{sec:supp_datasets} and \cref{sec:supp_implementation} respectively. More quantitative ablation study results are provided in \cref{sec:supp_ablation}. We further showcase the results for hierarchical segmentation, demonstrating the process of parsing any object into its semantic parts in \cref{sec:supp_visualization}. Finally, in \cref{sec:supp_prompt_mLLMs}, we explore the benefits of fine-grained prompts in images for mLLMs.

\section{Additional Evaluation Results}
\label{sec:supp_additional_results}
To illustrate the versatility and effectiveness of our model, we further compare the performance of our model with recent specialist and generalist models on object-level tasks, shown in \cref{table:object_level_tasks}. It turns out that our model achieves state-of-the-art performance on part-level tasks, while maintaining competitive performance on object-level tasks. This indicates that our model is capable of obtaining outstanding performance across tasks at different hierarchies, making it a foundation model that unifies both object-level and part-level tasks while acquiring multi-granularity recognition capabilities simultaneously.

\begin{table*}[!ht]
\caption{Comparison between \methodNAME with recent specialist and generalist models on object-level tasks. Note that for REC and RES tasks, we report Precision@0.5 and overall IoU (oIoU).}
\centering
\resizebox{1.0\linewidth}{!}{
\begin{tabular}{lccccccccccccccc} 
\toprule
\multirow{3}{*}{Method}   & \multirow{3}{*}{Type}     & \multicolumn{8}{c}{ {\it{Generic Detection \& Segmentation}}}     & \multicolumn{6}{c}{ {\it{Referring Detection \& Segmentation}}}      \\
 \cmidrule(lr){3-10} \cmidrule(lr){11-16}
 
&    & \multicolumn{2}{c}{COCO-val}  & \multicolumn{2}{c}{COCO-test-dev}    & \multicolumn{4}{c}{LVIS-val}   &  \multicolumn{2}{c}{RefCOCO}    & \multicolumn{2}{c}{RefCOCO+}      & \multicolumn{2}{c}{RefCOCOg} \\
 \cmidrule(lr){3-4}   \cmidrule(lr){5-6}   \cmidrule(lr){7-10}  \cmidrule(lr){11-12}  \cmidrule(lr){13-14}  \cmidrule(lr){15-16}
&  &$\rm AP_{box}$  & $\rm AP_{mask}$ &$\rm AP_{box}$  & $\rm AP_{mask}$  & $\rm AP_{box}$ & $\rm AP_{r-box}$  &$\rm AP_{mask}$  &$\rm AP_{r-mask} $ & $\rm P@0.5 $ &$\rm  oIoU$  & $\rm P@0.5 $ & $\rm oIoU $ & $\rm P@0.5 $ & $\rm oIoU $      \\ 
\hline
MDETR~\cite{mdetr}       & \multirow{8}{*}{Specialist}    & -   & -   & - & -   & -   & -   & -   & -     &87.5   & -   & 81.1   & -   & 83.4    & -    \\
SeqTR~\cite{seqtr}       &\multirow{8}{*}{Models}       & -   & -   & - & -   & -   & -   & -   & -    &87.0  &71.7  &78.7  &63.0  &82.7   & 64.7   \\
PolyFormer (L)~\cite{polyformer}  &   &-   & - & -   & -   & -   & -   & -   & -     &90.4   & 76.9   & 85.0   & 72.2   & 85.8    & 71.2 \\
ViTDet-L ~\cite{ViTDet}   &   &57.6  &49.8 &-   & -   &51.2  &-  &46.0  &34.3  & -   & -   & -   & -   & -   & -  \\
ViTDet-H ~\cite{ViTDet}  &   &58.7  &50.9 &-   & -   &53.4  &-  &48.1   &36.9  & -   & -   & -   & -   & -   & - \\
EVA-02-L~\cite{eva02} &    &64.2  &55.0   &64.5  &55.8 &65.2  &-   &57.3  &- &- & - &- &- &- &- \\
ODISE~\cite{ODISE}    &     & -   & - & -   & -   & -   & -   & -   & -     & -   & -   & -   & -   & -   & -   \\
Mask2Former (L)~\cite{mask2former}   &     & -   & 50.1   & -   & 50.5   & -   & -   & -   & -   & -     & -   & -   & -   & -   & -   \\
MaskDINO (L)~\cite{maskdino}   &     & -   & 54.5   & -   & 54.7    & -   & -   & -   & -   & -     & -   & -   & -   & -   &  - \\
\midrule
UniTAB (B)~\cite{unitab}           & \multirow{16}{*}{Generalist}   &-  & -  & -   & -  & -   & -   & -   & -     &88.6   & -   & 81.0   & -   & 84.6    & - \\ 
OFA (L)~\cite{OFA}  & \multirow{16}{*}{Models}  &-  & -   & -   & -   & -   & -   & -   & -     &90.1   & -   & 85.8   & -   & 85.9    & - \\
Pix2Seq v2 ~\cite{Pix2Seqv2}    &     &46.5   &38.2   & -   & -    & -   & -   & -   & -    & -   & -   &-   & -   & -   & -   \\ 
Uni-Perceiver-v2 (B)~\cite{Uni-perceiverv2}  &     &58.6   &50.6   & -   & - & -   & -   & -   & -     & -  & -   & -   & -   &-     & -   \\
Uni-Perceiver-v2 (L)~\cite{Uni-perceiverv2} &     &61.9  & 53.6  & -   & -    & -   & -   & -   & -     & -  & -   & -   & -   &-     & -   \\
UNINEXT (R50)~\cite{UNINEXT} &   &51.3   &44.9    & -   & -   &36.4   & -   & -   & -     & 89.7  & 77.9   & 79.8   & 66.2   &84.0    & 70.0   \\
UNINEXT (L)~\cite{UNINEXT} &     &58.1   &49.6   & -   & -   & -   & -   & -   & -     & 91.4   & 80.3   & 83.1   &70.0    & 86.9   & 73.4 \\
UNINEXT (H)~\cite{UNINEXT} &     &60.6   &51.8   & -   & -   & -   & -   & -   & -     & 92.6   & 82.2   & 85.2   &72.5    & 88.7  & 74.7  \\
GLIPv2 (B)~\cite{GLIPv2} &   &-   & - & 58.8   & 45.8  & -   & -   & -   & -   & -     & -   & -   & -   & -   & -   \\
GLIPv2 (H)~\cite{GLIPv2} &   &-   & - & 60.6   & 48.9  & -   & -   & -   & -   & -     & -   & -   & -   & -   & -   \\
X-Decoder (B)~\cite{xdecoder}  & &- &45.8 &- &45.8  & -   & -   & -   & -   & -   & -  & -   & -   & -  & -   \\
X-Decoder (L)~\cite{xdecoder}  & &- &46.7 &- &47.1   & -   & -   & -   & -   & -   & -  & -   & -   & -  & -  \\
Florence-2 (B)\cite{florence2}   &     &41.4   & -   & -   & -   & -   & -   & -    & -   &92.6   & -   & 86.8   & -   & 89.8   & -   \\
Florence-2 (L)~\cite{florence2}  &     &43.4   & -   & -  & -   & -   & -   & -   & -     &93.4   & -   & 88.3   & -   & 91.2 & - \\
GLEE (R50)\cite{wu2023GLEE}  &   &55.0   &48.4  &54.7  &48.3  &44.2  &36.7  &40.2  &33.7    &88.5   &77.4   &78.3    &64.8   &82.9   &68.8  \\
GLEE (L)\cite{wu2023GLEE} &   & 60.4   &53.0  &60.6 &53.3  &52.7  &44.5  &47.4  &40.4     &90.6   &79.5   &81.6    &68.3   &85.0   &70.6 \\
\hline
  \cellcolor{gray!20}\methodNAME(R50)
& \cellcolor{gray!20}Hierarchical
& \cellcolor{gray!20}54.4
& \cellcolor{gray!20}47.6
& \cellcolor{gray!20}54.2
& \cellcolor{gray!20}47.8
& \cellcolor{gray!20}42.7
& \cellcolor{gray!20}32.8  
& \cellcolor{gray!20}38.3  
& \cellcolor{gray!20}29.8 
& \cellcolor{gray!20}87.8   
& \cellcolor{gray!20}76.2  
& \cellcolor{gray!20}77.8    
& \cellcolor{gray!20}64.1   
& \cellcolor{gray!20}81.8   
& \cellcolor{gray!20}67.5  
\\
  \cellcolor{gray!20}\methodNAME(L)
& \cellcolor{gray!20}Models
& \cellcolor{gray!20}59.5
& \cellcolor{gray!20}52.0
& \cellcolor{gray!20}59.9
& \cellcolor{gray!20}52.5
& \cellcolor{gray!20}50.2
& \cellcolor{gray!20}39.6
& \cellcolor{gray!20}45.0
& \cellcolor{gray!20}35.9   
& \cellcolor{gray!20}89.6 
& \cellcolor{gray!20}78.4  
& \cellcolor{gray!20}80.3    
& \cellcolor{gray!20}67.2 
& \cellcolor{gray!20}84.0 
& \cellcolor{gray!20}69.5 
\\
\hline
\end{tabular}
}
\label{table:object_level_tasks}
\vspace{-2ex}
\end{table*}

\setlength{\tabcolsep}{4pt}
\begin{table}[t]
\caption{An ablation study on our Q-Former design. As we have demonstrated the effectiveness of independent decoders, we directly proceed this experiment starting from the structures of two independent decoders, with a backbone of ResNet-50.}
\begin{center}
\vspace{-1em}
\resizebox{0.8\columnwidth}{!}{
\begin{tabular}{clcccc}
\toprule
\multirow{2}{*}{Training Datasets} &\multirow{2}{*}{Method} &\multicolumn{2}{c}{$\rm bbox$} & \multicolumn{2}{c}{$\rm segm$}\\ 
\cmidrule(lr){3-4} \cmidrule(lr){5-6}
& &\textit{$\rm AP$} &\textit{$\rm AP_{50}$} &\textit{$\rm AP$} &\textit{$\rm AP_{50}$} \\
\noalign{\smallskip}
\hline
\noalign{\smallskip}
\multirow{3}{*}{Pascal Part Base $+$ VOC}
& Independent Decoders& 6.7 & 12.3 & 5.6 & 11.2  \\ 
& $+$ Q-Former & 8.0 & 14.3 & 6.9 & 13.6 \\
& \textcolor{darkgreen}{\emph{vs. baseline}}&\textcolor{darkgreen}{+1.3} &\textcolor{darkgreen}{+2.0}&\textcolor{darkgreen}{+1.3}&\textcolor{darkgreen}{+2.4} \\
\bottomrule
\end{tabular}}
\end{center}
\label{table:ablation_QFormer}
\vspace{-2em}
\end{table}
\setlength{\tabcolsep}{1.4pt}
\setlength{\tabcolsep}{4pt}
\begin{table}[t]
\caption{An ablation study on $topK$ object queries and $L$ universal parsing queries. We directly conduct this experiment with a backbone of ResNet-50 and train our model on PACO dataset for 90K iterations.}
\begin{center}
\vspace{-1em}
\resizebox{0.7\columnwidth}{!}{
\begin{tabular}{ccccccccc}
\toprule
\multirow{2}{*}{Id} & \multirow{2}{*}{$topK$} & \multirow{2}{*}{L} & \multicolumn{6}{c}{PACO} \\
\cmidrule(lr){4-9}
& & & $\rm AP_{box}$ & $\rm AP_{box}^{obj}$ & $\rm AP_{box}^{opart}$ & $\rm AP_{mask}$ & $\rm AP_{mask}^{obj}$ & $\rm AP_{mask}^{opart}$ \\
\noalign{\smallskip}
\hline
\noalign{\smallskip}
$1$ & 50 & 10 & 29.4 & 52.7 & 22.0 & 23.1 & 48.1 & 16.8 \\ 
$2$ & 50 & 20 & 29.7 & 53.7 & 22.4 & 23.4 & 48.7 & 16.9 \\
$3$ & 50 & 30 & 29.4 & 52.7 & 22.4 & 23.2 & 48.0 & 16.9 \\
$4$ & 75 & 10 & 29.0 & 53.0 & 22.0 & 23.1 & 48.4 & 17.0 \\
$5$ & 100 & 10 & 29.7 & 52.9 & 22.8 & 23.3 & 48.0 & 17.4 \\
\bottomrule
\end{tabular}}
\end{center}
\label{table:ablation_TopK_L}
\end{table}
\setlength{\tabcolsep}{1.4pt}

\section{Datasets Unification}
\label{sec:supp_datasets}

To facilitate the training process of the Q-Former, we augment the original part-level datasets with object-level annotations to establish hierarchical correspondences. Specifically, we add object-level annotations to Pascal Part\cite{chen2014detect}, PartImageNet\cite{he2021partimagenet}, Pascal-Part-116\cite{wei2023ov}, ADE-Part-234\cite{wei2023ov}, in order to establish the hierarchical correspondence between objects and parts. 
It is necessary to clarify that both Pascal-Part-116 and ADE-Part-234 only provide semantic segmentation annotations, which cannot be directly employed for joint-training. Consequently, we utilize the erosion function from the skimage library to convert them into instance segmentation annotations.
Besides, we note that Pascal-Part-116 offers more part-level annotations than Pascal Part, with a relatively larger amount of images. Some images are shared between the two datasets. 

\setlength{\tabcolsep}{4pt}
\begin{table}[t]
\caption{\textbf{Pascal Part part taxonomy} from \cite{peize2023vlpart}.}
\begin{center}
\vspace{-2em}
\resizebox{0.8\columnwidth}{!}{
\begin{tabular}{c r l l l}
\toprule
Dataset Name & Id & Type & Object Categories &  Part Categories\\
\midrule
\multirow{20}{*}{Pascal Part\cite{chen2014detect}}
& 1 & Base & aeroplane & body, wing, tail, wheel \\
& 2 & Base & bicycle & wheel, handlebar, saddle \\
& 3 & Base & bird & beak, head, eye, foot, leg, wing, neck, tail, torso \\
& 4 & Base & boat & - \\
& 5 & Base & bottle & body, cap \\
& 6 & Novel & bus & license plate, door, headlight, mirror, window, wheel\\
& 7 & Base & car & license plate, door, headlight, mirror, window, wheel\\
& 8 & Base & cat & head, leg, paw, ear, eye, neck, nose, tail, torso \\
& 9 & - & chair & - \\
& 10 & Base & cow & head, leg, ear, eye, horn, muzzle, neck, tail, torso\\
& 11 & - & diningtable & -\\
& 12 & Novel & dog & head, leg, paw, ear, eye, muzzle, neck, nose, tail, torso \\
& 13 & Base & horse & head, leg, ear, eye, muzzle, neck, tail, torso \\
& 14 & Base & motorbike & wheel, handlebar, headlight, saddle \\ 
& 15 & Base & person & hair, head, ear, eye, nose, neck, mouth, arm, hand, leg, foot, torso \\
& 16 & Base &  pottedplant & plant, pot\\
& 17 & Base & sheep & head, leg, ear, eye, horn, muzzle, neck, tail, torso \\
& 18 & - & sofa & -\\
& 19 & - & train & -\\
& 20 & - & tvmonitor & -\\
\bottomrule
\end{tabular}}
\end{center}
\label{table:pascalpart_categories}
\vspace{-2em}
\end{table}
\setlength{\tabcolsep}{1.4pt}
\textbf{Pascal Part.} We utilize the modified version of Pascal Part provided by VLPart\cite{peize2023vlpart}, which contains 93 part-level categories in total. The Object-level annotations are directly acquired from the VOC dataset and are then integrated with the original part-level annotations to establish hierarchical correspondences. The semantic parts of bus and dog are selected as the novel parts, totally 16 parts, the remaining 77 parts are base categories, as shown in \cref{table:pascalpart_categories}.

\setlength{\tabcolsep}{4pt}
\begin{table}[t]
\caption{\textbf{PartImageNet part taxonomy} from \cite{he2021partimagenet}.}
\begin{center}
\vspace{-2em}
\resizebox{0.7\columnwidth}{!}{
\begin{tabular}{c r l l}
\toprule
Dataset Name & Id & Object Categories &  Part Categories\\
\midrule
\multirow{11}{*}{PartImageNet\cite{he2021partimagenet}}
& 1 & Quadruped & head, body, foot, tail \\
& 2 & Biped & head, body, hand, foot, tail \\
& 3 & Fish & head, body, fin, tail\\
& 4 & Bird & head, body, wing, foot, tail \\
& 5 & Snake & head, body \\
& 6 & Reptile & head, body, foot, tail\\
& 7 & Car & body, tier, side mirror \\
& 8 & Bicycle & head, body, seat, tier \\
& 9 & Boat & body, sail \\
& 10 & Aeroplane & head, body, wing, engine, tail \\
& 11 & Bottle & body, mouth\\
\bottomrule
\end{tabular}}
\end{center}
\label{table:partimagenet_categories}
\vspace{-2em}
\end{table}
\setlength{\tabcolsep}{1.4pt}
\textbf{PartImageNet.} PartImageNet selects 158 classes from the ImageNet dataset and organize them into 11 super-categories. Since each image in this PartImageNet only contains part-level annotations for an individual object, we simply merge the bounding boxes and masks from these part-level annotations to derive object-level annotations.
{For boxes, we create a single largest bounding box as object box to encloses all part boxes. For masks, we take the union of all part masks as the object masks.}
The detailed categoreis are listed in \cref{table:partimagenet_categories}.

\setlength{\tabcolsep}{4pt}
\begin{table}[t]
\caption{\textbf{Pascal-Part-116 part taxonomy} from \cite{wei2023ov}.}
\begin{center}
\vspace{-1em}
\resizebox{0.9\columnwidth}{!}{
\begin{tabular}{c r l l l}
\toprule
Dataset Name & Id & Type & Object Categories &  Part Categories\\
\midrule
\multirow{20}{*}{Pascal-Part-116\cite{wei2023ov}}
& 1 & Base & aeroplane & body, stern, wing, tail, engine, wheel \\
& 2 & Base & bicycle & wheel, saddle, handlebar, chainwheel, headlight \\
& 3 & Novel & bird & wing, tail, head, eye, beak, torso, neck, leg, foot \\
& 4 & Base & boat & - \\
& 5 & Base & bottle & body, cap \\
& 6 & Base & bus & wheel, headlight, front, side, back, roof,
mirror, license plate, door, window\\
& 7 & Novel & car & wheel, headlight, front, side, back, roof,
mirror, license plate, door, window]\\
& 8 & Base & cat & tail, head, eye, torso, neck, leg, nose, paw, ear \\
& 9 & - & chair & - \\
& 10 & Base & cow & tail, head, eye, torso, neck, leg, ear, muzzle, horn\\
& 11 & - & diningtable & -\\
& 12 & Novel & dog & tail, head, eye, torso, neck, leg, nose,
paw, ear, muzzle \\
& 13 & Base & horse & tail, head, eye, torso, neck, leg, ear, muzzle, hoof \\
& 14 & Novel & motorbike & wheel, saddle, handlebar, headligh \\ 
& 15 & Base & person & head, eye, torso, neck,
leg, foot, nose, ear, eyebrow, mouth, hair, lower arm, upper arm, hand \\
& 16 & Base & pottedplant & pot, plant\\
& 17 & Novel & sheep & tail, head, eye, torso, neck, leg, ear, muzzle, horn] \\
& 18 & - & sofa & -\\
& 19 & Base & train & headlight, head, front, side, back, roof, coach\\
& 20 & Base & tvmonitor & screen\\
\bottomrule
\end{tabular}}
\end{center}
\label{table:pascal_part_116_categories}
\vspace{-2em}
\end{table}
\setlength{\tabcolsep}{1.4pt}
\textbf{Pascal-Part-116.} Similar to Pascal Part, we also acquire object-level annotations from the VOC dataset on the corresponding images, integrating them with the original part-level annotations. The semantic parts of bird, car, dog, motorbike and sheep are selected as novel categories, as shown in \cref{table:pascal_part_116_categories}.

\setlength{\tabcolsep}{4pt}
\begin{table}[t]
\caption{Example phrases at different hierarchies in Visual Genome.}
\begin{center}
\vspace{-1.5em}
\resizebox{0.98\columnwidth}{!}{
\begin{tabular}{c l}
\toprule
Hierarchy &  Phrases\\
\midrule
\multirow{5}{*}{Object-level}
& fruits, piping, traffic light, suit jacket, chipmunk, bee, sidewalk curb, golf clubs, tennis, dog, yellow fire hydrant, \\
& kite, romaine, stone landscape, man and women, iced tea, this is a cow, child hotdog, television, smartphone, swimmers,\\
& green watermelon, flat ground, green grapes, demon, large picture, cucumber pile, white curl, seven arched windows, door/wall, \\
& left jean pant, business place, snow resort, alliance, coffee saucer, bench swing, emperor, sliced gourds, purple toboggan, orange petals, \\
& squared shirt, lighting system, lamb standing, broadcaster, green wallet, tourist attraction, iced donuts, blue bench, paper napkin ···\\
\midrule
\multirow{6}{*}{Part-level} 
& cat whisker, pilot's seat, baby's mouth, porcelain tile, cap head, giraffe head, part of the sky, cow's neck, person's mid finger \\
& bus front, laptop mouse, door plate, plane's tail, handle on teapot, shadow of scooter, nail on finger, players foot, finger pointing, metal lightpole\\
& set of wheels, player's waist, wood grained, bronze knob, bar handle, pajama top, button labels, crosswalk button, balcony rail, sheep skin\\
& hand is on clock, door lock, sink edge, woman's top, keys on the keychain, ear flap, cow legs, rubber foot, knee support, heating plate\\
& photo red eye, blue tank, front fender, lace collar, silver hand rail, leg is yellow, windows of plane, blue reins, tusk, sheep's hair\\
& brown eye, small/black wheel, swing arm, cordless mouse, log leg, brown eye, face guard, plane's propeller, tiger's eye, train windshield···\\
\bottomrule
\end{tabular}}
\end{center}
\label{table:vg_phrases}
\vspace{-2em}
\end{table}
\setlength{\tabcolsep}{1.4pt}

\setlength{\tabcolsep}{4pt}
\begin{table}[t]
\caption{\textbf{ADE-Part-234 part taxonomy} from \cite{wei2023ov}.}
\begin{center}
\vspace{-2em}
\resizebox{0.9\columnwidth}{!}{
\begin{tabular}{c r l l l}
\toprule
Dataset Name & Id & Type & Object Categories &  Part Categories\\
\midrule
\multirow{44}{*}{ADE-Part-234\cite{wei2023ov}}
& 1 & Base & person & arm, back, foot, gaze, hand, head, leg, neck, torso \\
& 2 & Base & door & door frame, handle, knob, panel \\
& 3 & Base & clock & face, frame \\
& 4 & Base & toilet & bowl, cistern, lid \\
& 5 & Base & cabinet & door, drawer, front, shelf, side, skirt, top \\
& 6 & Base & sink & bowl, faucet, pedestal, tap, top \\
& 7 & Base & lamp & arm, base, canopy, column, cord, highlight,light source, shade,tube \\
& 8 & Base & sconce & arm, backplate, highlight, light source, shade \\
& 9 & Base & chair & apron, arm, back, base, leg, seat, seat cushion, skirt, stretcher \\
& 10 & Base & chest of drawers & apron, door, drawer, front, leg \\
& 11 & Base & chandelier & arm, bulb, canopy, chain, cord, highlight, light source, shade \\
& 12 & Base & bed & footboard, headboard, leg, side rail \\
& 13 & Base & table & apron, drawer, leg, shelf, top, wheel\\
& 14 & Base & armchair & apron, arm, back, back pillow, leg, seat, seat base, seat cushion \\
& 15 & Novel & ottoman & back, leg, seat \\
& 16 & Base & shelf & door, drawer, front, shelf \\
& 17 & Novel & swivel chair & back, base, seat, wheel \\
& 18 & Novel & fan & blade, canopy, tube \\
& 19 & Base & coffee table & leg, top\\
& 20 & Novel & stool & leg, seat \\
& 21 & Base & sofa & arm, back, back pillow, leg, seat base, seat cushion, skirt  \\
& 22 & Base & computer & computer case, keyboard, monitor, mouse \\
& 23 & Novel & desk & apron, door, drawer, leg, shelf, top \\
& 24 & Base & wardrobe & door, drawer, front, leg, mirror, top \\
& 25 & Base & car & bumper, door, headlight, hood, license plate, logo, mirror, wheel, window, wiper \\
& 26 & Novel & bus & bumper, door, headlight, license plate, logo, mirror, wheel, window, wiper \\
& 27 & Novel & oven & button panel, door, drawer, top \\
& 28 & Base & cooking stove & burner, button panel, door, drawer, oven, stove\\
& 29 & Base & microwave & button panel, door, front, side, top, window \\
& 30 & Base & refrigerator & button panel, door, drawer, side \\
& 31 & Novel & kitchen island & door, drawer, front, side, top \\
& 32 & Base & dishwasher & button panel, handle, skirt \\
& 33 & Base & bookcase & door, drawer, front, side\\
& 34 & Base & television receiver & base, buttons, frame, keys, screen, speaker \\
& 35 & Base & glass & base, bowl, opening, stem \\
& 36 & Base & pool table & bed, leg, pocket \\
& 37 & Novel & van & bumper, door, headlight, license plate, logo, mirror, taillight, wheel, window, wiper \\
& 38 & Base & airplane & door, fuselage, landing gear, propeller, stabilizer, turbine engine, wing \\
& 39 & Novel & truck & bumper, door, headlight, license plate, logo, mirror, wheel, windshield\\
& 40 & Novel & minibike & license plate, mirror, seat, wheel \\
& 41 & Base & washer & button panel, door, front, side \\
& 42 & Novel & bench & arm, back, leg, seat \\
& 43 & Base & traffic light & housing, pole \\
& 44 & Base & light & aperture, canopy, diffusor, highlight, light source, shade \\
\bottomrule
\end{tabular}}
\end{center}
\label{table:ade_part_234_categories}
\vspace{-3em}
\end{table}
\setlength{\tabcolsep}{1.4pt}
\textbf{ADE-Part-234.} We obtain object-level annotations from the ADE20K-Instance dataset for ADE-Part-234. Throughout the process, we solely utilize the object-level annotations corresponding to the images in ADE-Part-234, without introducing additional object categories from ADE20K dataset. The detailed categorization of base and novel classes is presented in \cref{table:ade_part_234_categories}.

\textbf{PACO.} PACO contains 75 object categories and 456 object
part categories, as shown in \cref{table:paco_categories}. Note that the PACO dataset includes annotations for both objects and parts, obviating the need for any modifications. Consequently, we directly utilize the original annotations for joint-training.

\begin{figure*}[tb]
\centering
\includegraphics[width=0.95 \linewidth]{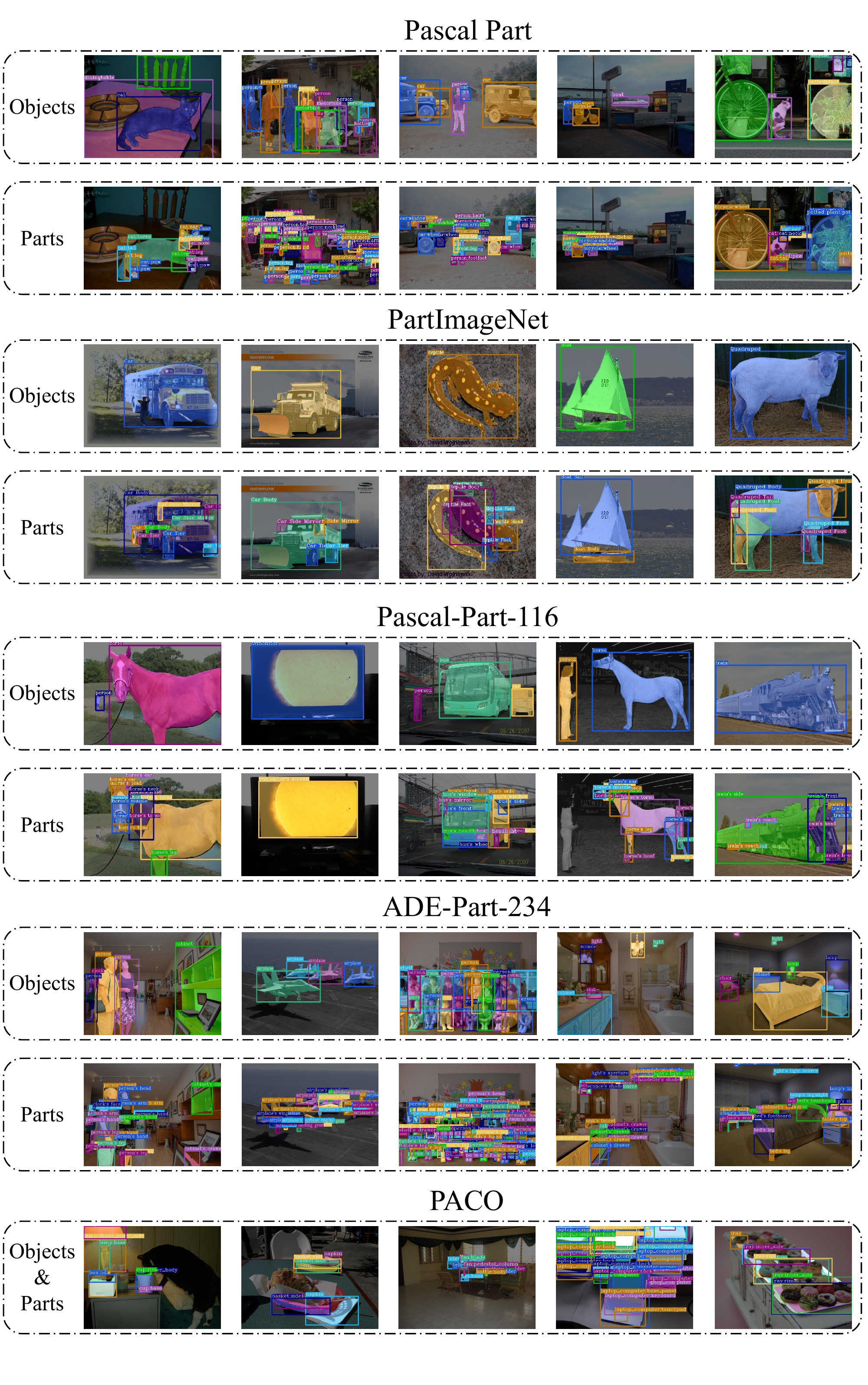}
\caption{
\textbf{Visualization of hierarchical correspondences in part-level datasets.}
}
\label{fig:hierarchical_data}
\vspace{-2ex}
\end{figure*}

Hence, the annotation granularity of part-level datasets are standardized by adding object-level annotations, complementing these datasets with hierarchical correspondences. The visualization of hierarchical correspondences between objects and parts are illustrated in \cref{fig:hierarchical_data}.

To further improve the generalization capability of our Q-Former, we organize Visual Genome\cite{visualgenome} and SA-1B\cite{SAM} into hierarchical versions for joint-training.

\textbf{Visual Genome.} Since Visual Genome contains multiple instances on a single image, we treat it as a detection task and divide its noun phrases into object level and part level. Eventually, the Visual Genome dataset is annotated with both object and part hierarchies, including 45,054 object-level phrases and 25,109 part-level phrases. We display some object-level and part-level phrases from the VG dataset in \cref{table:vg_phrases}.

\begin{figure*}[tb]
\centering
\includegraphics[width=0.95 \linewidth]{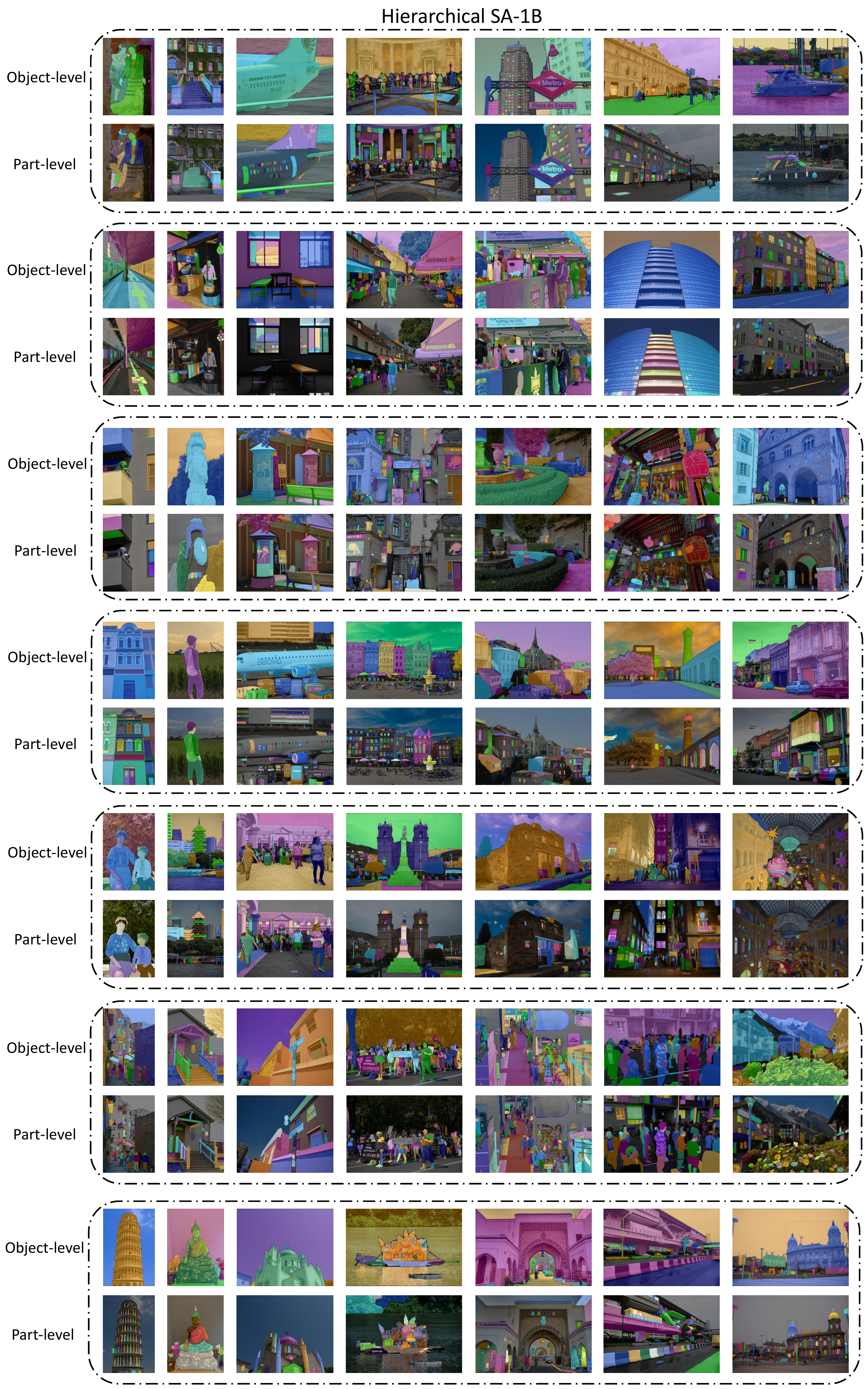}
\caption{
\textbf{Visualization of our proposed hierarchical SA-1B.}
}
\label{fig:hierarchical_SA_1B}
\vspace{-2ex}
\end{figure*}

\textbf{SA-1B.} We introduce a subset of the open-world instance segmentation dataset SA-1B\cite{SAM} to further improve the generalization capability of our model. As SA-1B provides abundant class-agnostic mask annotations and the Segment Anything Model (SAM) is able to perform multi-level segmentation, we observe that many masks exhibit a certain degree of overlap with each other, indicating a granularity distinction among masks. Therefore, we propose to calculate the overlap ratio $R$ between pairs of masks and set a threshold to identify masks at different granularities. This process can be denoted as:
\begin{align}
    \vspace{-2ex}
    R_{ij} = \frac{\lvert S_i \cap S_j \rvert}{max(S_i, S_j)}
    \label{eq:8}
    \vspace{-2ex}
\end{align}
where $S_i$, $S_j$ represents the area of $i-th$ mask and $j-th$ mask respectively, and $R_{ij}$ stands for the overlap ratio between the $i-th$ mask and the $j-th$ mask. We propose to compute the overlap ratio $R$ by adopting a division operation between the intersection area of two masks and the area of the larger mask. When the overlap ratio $R$ between two masks exceeds the threshold $t$, we consider the mask with a larger area as an object-level mask, while the other one with a smaller area is classified as a part-level mask. In our experiment, we set the threshold $t=0.5$ and convert a subset of SA-1B into a hierarchical dataset. During the training process, we set the category name for each instance to be ‘object’ or ‘part’ in accordance with the hierarchy of its corresponding annotation. We then perform joint-training with our hierarchical SA-1B in instance segmentation paradigm. The visualization of our proposed hierarchical SA-1B are shown in \cref{fig:hierarchical_SA_1B}.

\section{Implementation Details}
\label{sec:supp_implementation}

Following \cite{maskdino, wu2023GLEE}, we utilize a image backbone, a text encoder, a 6-layer deformable transformer encoder for pixel decoding and two independent 9-layer decoders for generating hierarchical predictions. We adopt $300$ object queries and $10$ universal parsing queries throughout our experiments. We observe that directly combining MaskDINO and CLIP\cite{CLIP} and training from scratch  will result in exceedingly difficult convergence. Thus, unless otherwise specified, our model is initialized with the pre-trained weight from GLEE on Object365 and OpenImages, both of which are object-level datasets, and the CLIP text encoder uses the frozen original weights. We use AdamW\cite{AdamW} optimizer with a base learning rate of $5\times10^{-5}$ and a weight decay of $0.05$ at the $12,000$ iterations and $16,000$ iterations by a factor of 0.1 for training on zero-shot part segmentation task. As for joint-training, we directly load the weight of GLEE and continue training for $200,000$ iterations, the learning rate of the image backbone is multiplied by a factor of $0.1$.
Through experiments, we notice that if we perform joint-training with a unified matching process instead of a decoupled matching mechanism, our independent decoders will confuse objects and parts, thereby generating similar predictions. Consequently, we introduce a decoupled matching mechanism following \cite{wang2024hierarchical}, encouraging the independent decoders at different hierarchies to learn distinctive features associated with objects or parts respectively.

\section{Ablation Study}
\label{sec:supp_ablation}

\textbf{Ablation on Q-Former.}
To demonstrate the effectiveness of our Q-Former, we conduct an ablation experiment on this structure. As we have validated that two independent decoders achieves a favorable outcome, our experiment directly starts from this configuration, comparing the performance with and without the Q-Former design. Our experimental setup involves a joint training on the Pascal Part Base and VOC datasets for 5000 iterations, followed by a zero-shot evaluation on the PartImageNet dataset. The inclusion of the VOC dataset aims to provide our model with additional hierarchical information about objects, thereby examining its capability to effectively transfer this knowledge to the part level. The results are shown in \cref{table:ablation_QFormer}. It turns out that incorporating the Q-Former indeed facilitates the effective transfer of hierarchical information from objects to parts, thereby improving the performance of our model.

\textbf{Ablation on Hyperparameters.}
In this ablation study, we employ a ResNet-50 backbone and train our model on PACO for 90K iterations. We observe that increasing $topK$ and $L$ may result in a slight improvement in $\rm AP^{opart}$. 
However, augmenting $topK$ and $L$ incurs additional training time and GPU memory usage without yielding commensurate performance gains. Thus, we set $topK=50$ and $L=10$ as a trade-off configuration to ensure training efficiency.

\begin{figure}[ht]
    \centering
    \vspace{-2ex}
    \includegraphics[width=0.7 \linewidth]{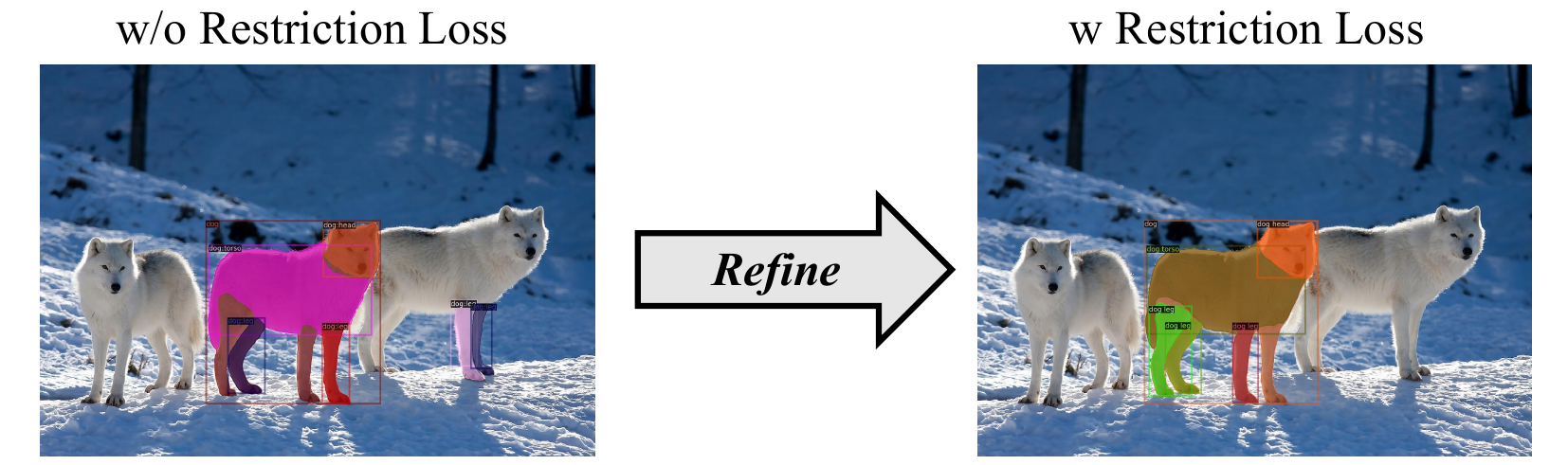}
    \caption{\textbf{Visualization of the effectiveness after adopting the Restriction Loss.}}
    \label{fig:restriction_loss}
    \vspace{-6ex}
\end{figure}

\textbf{Effectiveness of Box Restriction Loss.}
As PACO treats objects and parts as distinct instances during the annotation process, it is not guaranteed that an object in an image will necessarily have its corresponding parts, and vice versa. Consequently, during the training process, it is highly probable that the part queries corresponding to a certain object may be matched with the part annotations of another object nearby, leading to the confusion of the hierarchical relationships. To constrain the mutual correspondence between an object and its constituent parts, we introduce a Restriction Loss $L_{res}$, which penalizes those part prediction boxes that extend beyond the corresponding object prediction box. As shown in \cref{fig:restriction_loss}, we visualize an object prediction with a highest confidence score alongside their corresponding highly confident part predictions. It turns out that after incorporating the Restriction Loss on boxes, the predicted boxes are indeed constrained, eliminating the occurrences of part predictions drifting towards parts of another object.

\section{Visualization Results}
\label{sec:supp_visualization}

\begin{figure}[H]
\centering
\vspace{-4ex}
\includegraphics[width=0.95 \linewidth]{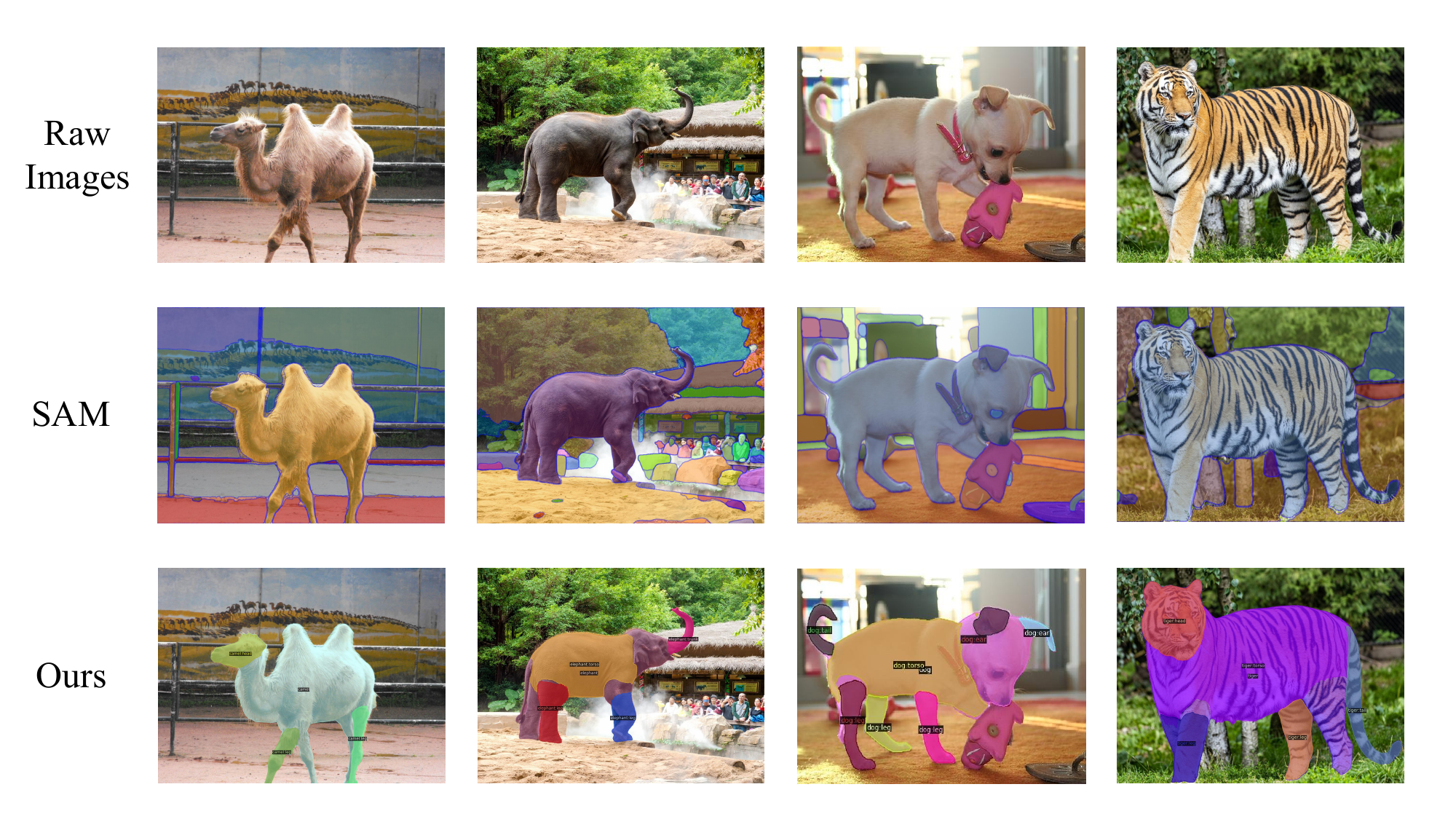}
\caption{Comparison of visualization results between SAM\cite{SAM} and \methodNAME.
}
\label{fig:SAM_comparison}
\vspace{-4ex}
\end{figure}

\textbf{Comparison with Segment Anything Model.} Segment Anything Model (SAM)\cite{SAM} is a prompt-based model designed for performing interactive segmentation tasks. It is capable of generating masks across multiple granularities including both objects and their parts in an image. As shown in \cref{fig:SAM_comparison}, when directly comparing the visualization results of SAM and \methodNAME, we observe that SAM, being a class-agnostic segmentation model, predominantly relies on features such as colors or boundaries of instances within the image for segmentation. As a result, it faces challenges in distinguishing various components of a certain object especially for animals. In contrast, \methodNAME exhibits the capability to comprehend the semantics of objects as well as their respective parts. Consequently, our model is able to recognize and parse any object to obtain its corresponding parts.

\begin{figure}[H]
\centering
\vspace{-4ex}
\includegraphics[width=0.95 \linewidth]{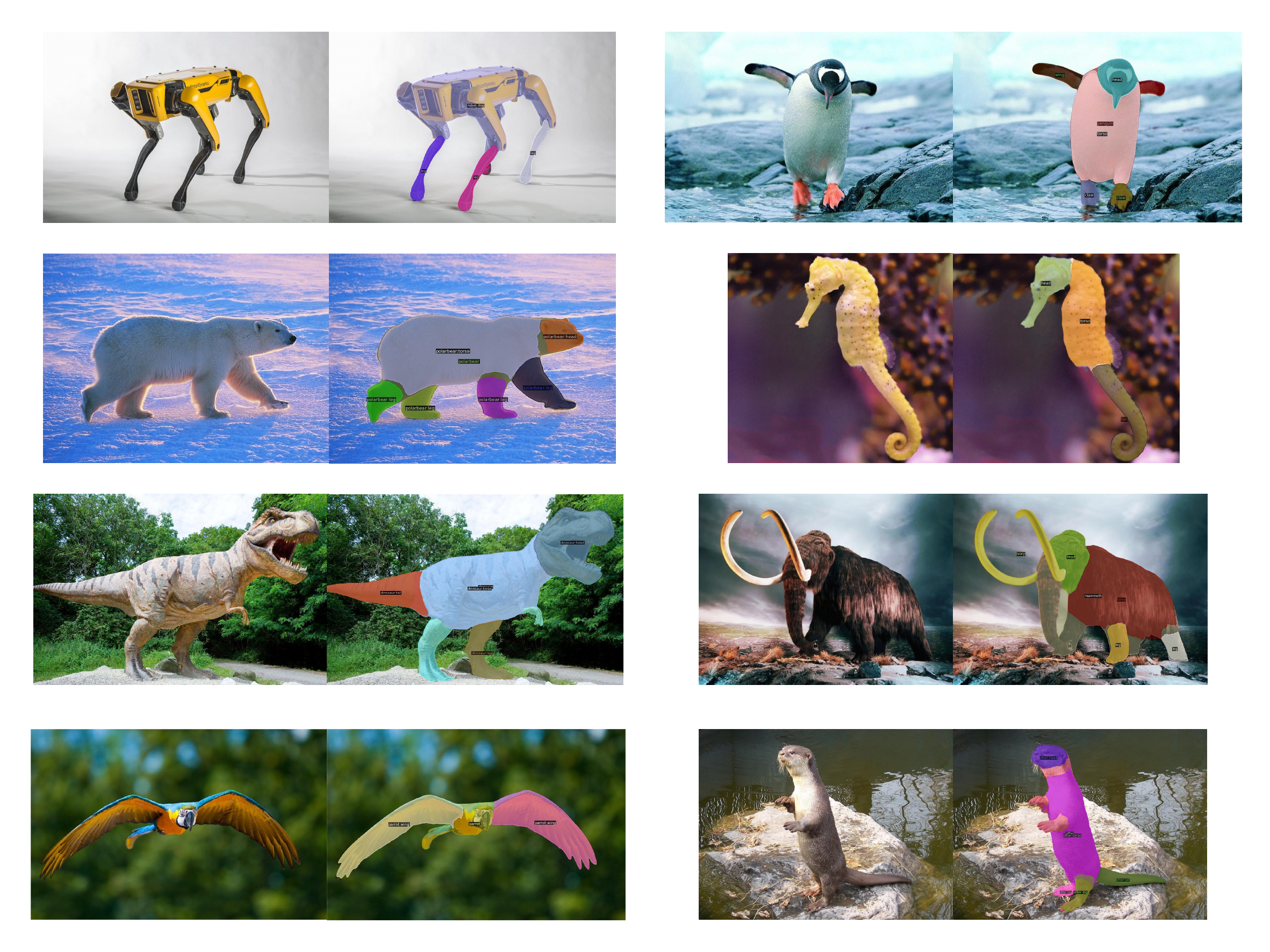}
\caption{Visualization of the generalization capability of \methodNAME.
}
\label{fig:generalization_visualization}
\vspace{-4ex}
\end{figure}

\textbf{Visualization of the Generalization Capability.}
We evaluate the generalization ability of our model on novel objects. By establishing a hierarchical relationship between objects and parts via Q-Former, our model exhibits strong generalization capability, which enables it to first recognize and then parse these objects into their corresponding semantic parts, as illustrated in \cref{fig:generalization_visualization}. Among the evaluated categories, robot-dog, penguin, polarbear, hippocampus, dinosaur, mammoth, parrot, and otter have never been encountered in the part-level datasets; yet, their parts are still segmented accurately, demonstrating the robust generalization of our method to novel objects. 

\begin{figure}[tb]
\centering
\includegraphics[width=1.0 \linewidth]{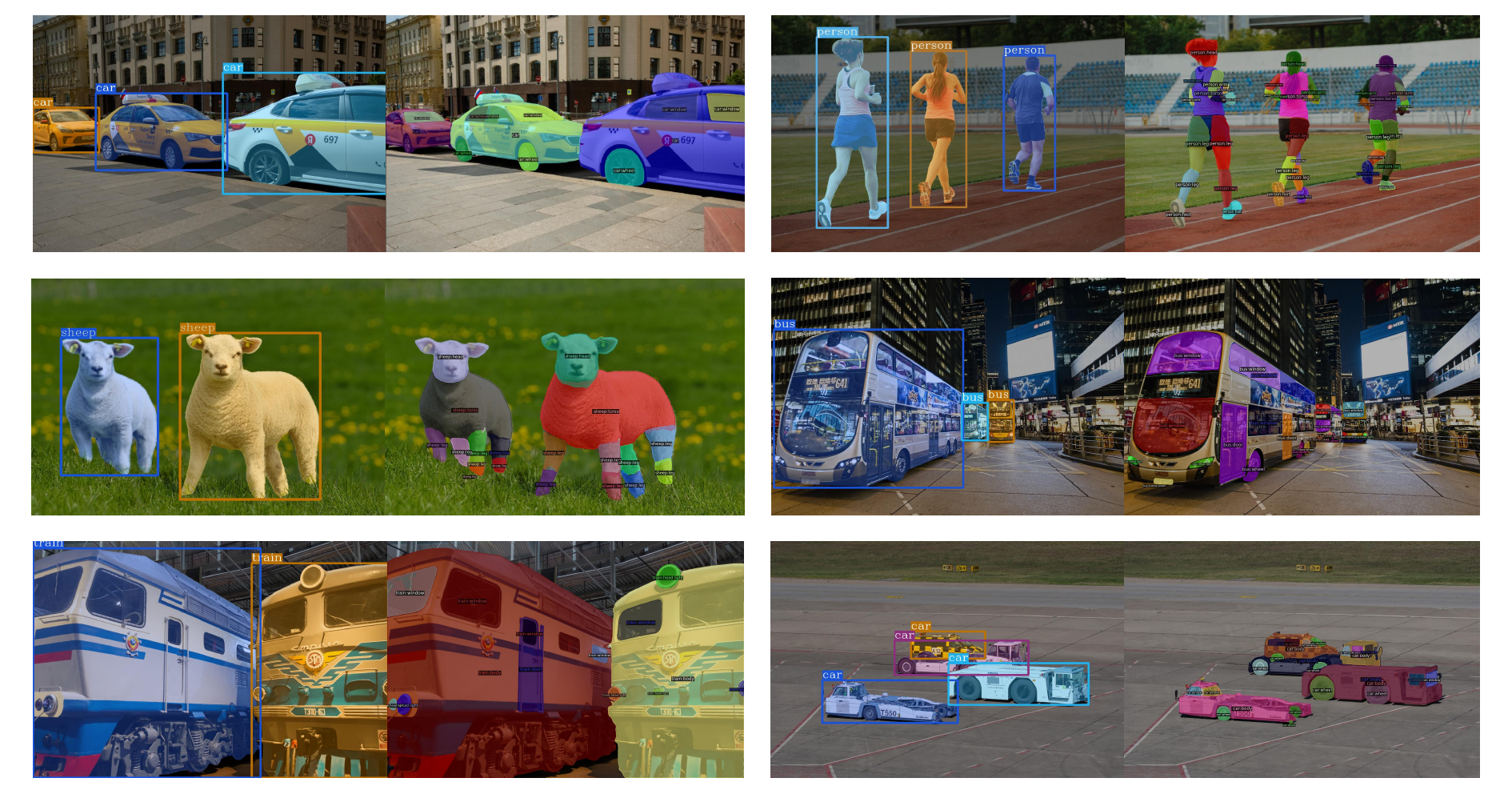}
\caption{Visualization of parsing objects into their corresponding parts.
}
\label{fig:parse_object_into_parts}
\end{figure}

\begin{figure}[tb]
\centering
\includegraphics[width=1.0 \linewidth]{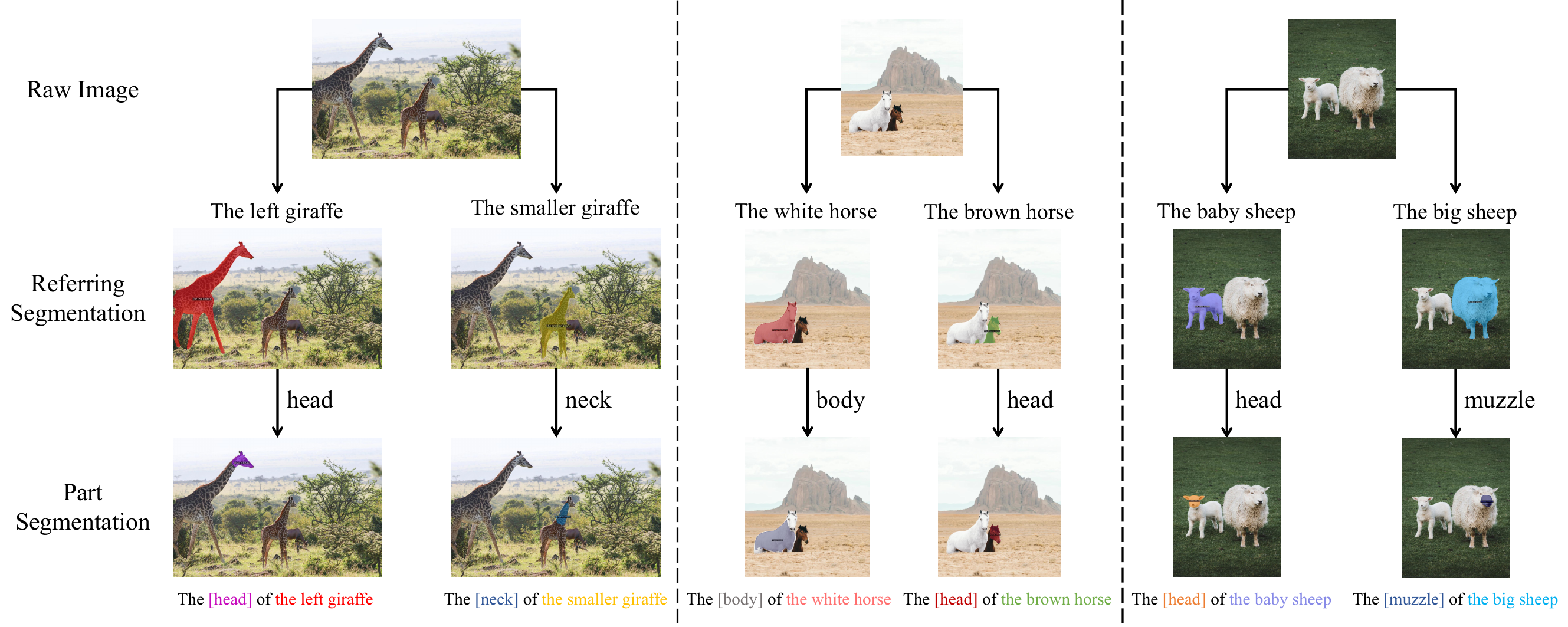}
\caption{Visualization of referring part segmentation.
}
\label{fig:referring_visualization}
\vspace{-2ex}
\end{figure}

\textbf{Parse Any Object into Parts.}
\cref{fig:parse_object_into_parts} demonstrates the capability of our model to accurately parse each object into its corresponding parts in daily scenes. Furthermore, by leveraging a vast amount of region-level expression data for joint training following GLEE~\cite{wu2023GLEE}, the object decomposition ability can be generalized to objects referred by expressions. 
For parts referred by any linguistic expression, we first identify the corresponding object and its object query by matching with the expression. The Q-Former parses the object query into corresponding part queries, which then facilitate the retrieval of the relevant part, achieving the ability to parse any object into parts with a detailed expression,
shown in \cref{fig:referring_visualization}.

\section{Experiment on mLLMs}
\label{sec:supp_prompt_mLLMs}

\begin{figure*}[tb]
\centering
\includegraphics[width=0.9 \linewidth]{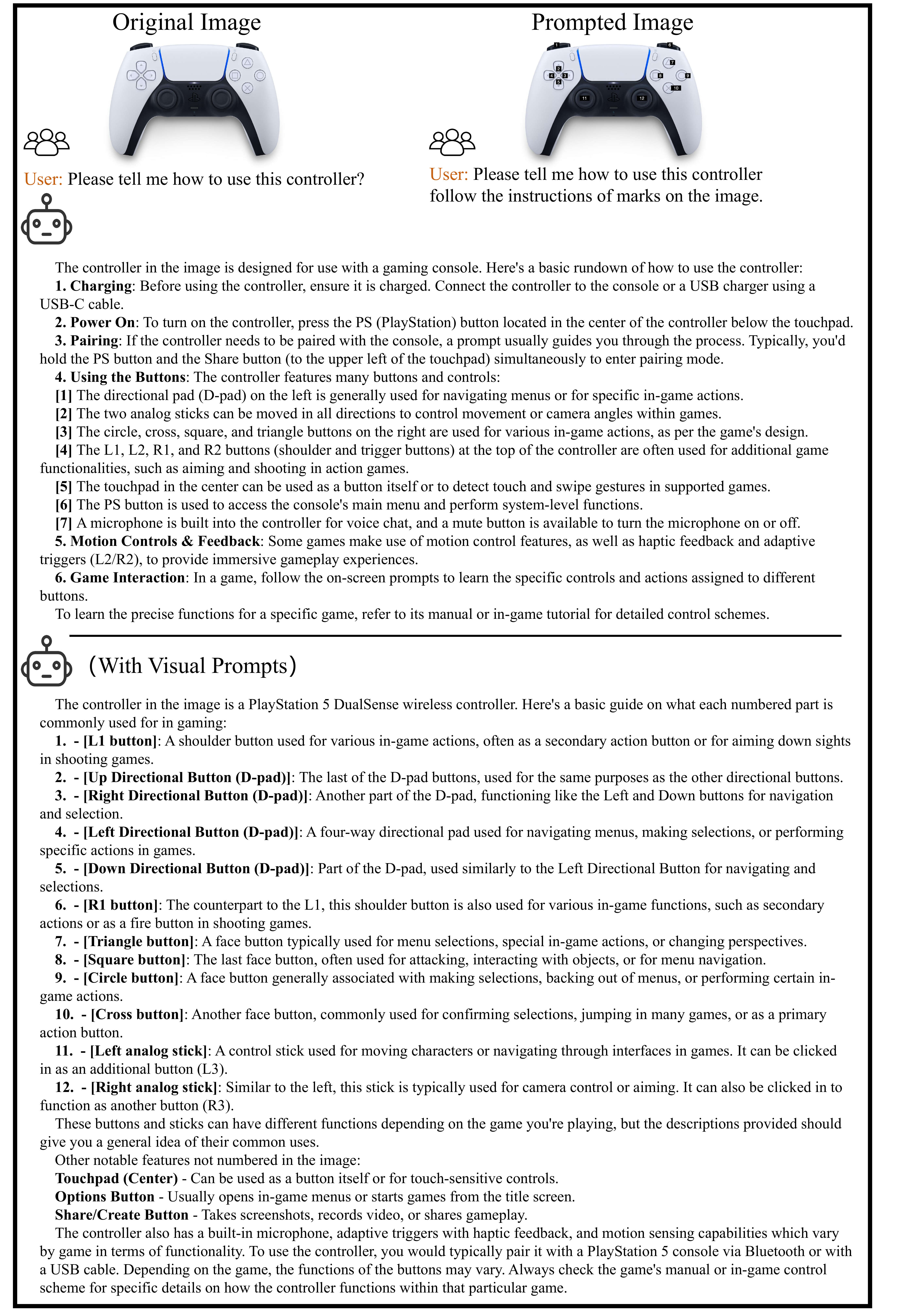}
\caption{
GPT4V with fine-grained visual prompts can help to provide detailed information on tool usage, teaching users the function of each button on a controller. We observe that when there are no specific visual prompts in the image, GPT4V may output some sentences that are not directly related to the image content, such as "charging", "power on", "pairing". Prompting GPT4V with a set of marks enables it to have a detailed comprehension of the information within the image, thereby generating more satisfactory answers.
}
\label{fig:gpt4v_ps5}
\vspace{-2ex}
\end{figure*}

\begin{figure*}[tb]
\centering
\includegraphics[width=0.99 \linewidth]{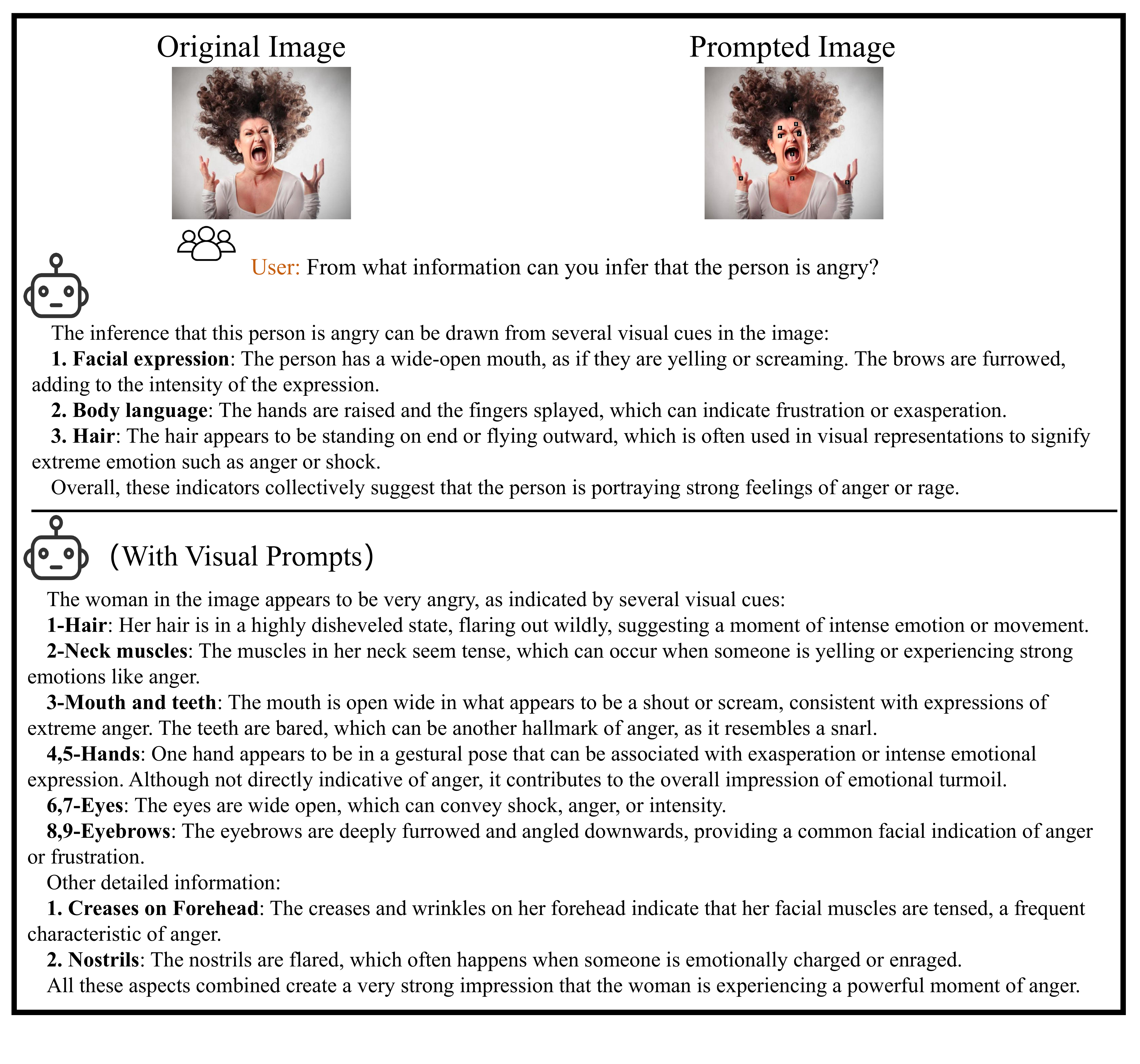}
\caption{
Providing GPT4V with detailed visual prompts improves its ability to acquire further insight and comprehension of the image. In instance where visual prompts are absent in the image, GPT4V may only provide general responses. However, when adding visual prompts on the image, GPT4V can effectively respond to specific details and offer answers that are more aligned with the content of the image.
}
\label{fig:gpt4v_face}
\vspace{-2ex}
\end{figure*}

\begin{figure*}[tb]
\centering
\includegraphics[width=0.99 \linewidth]{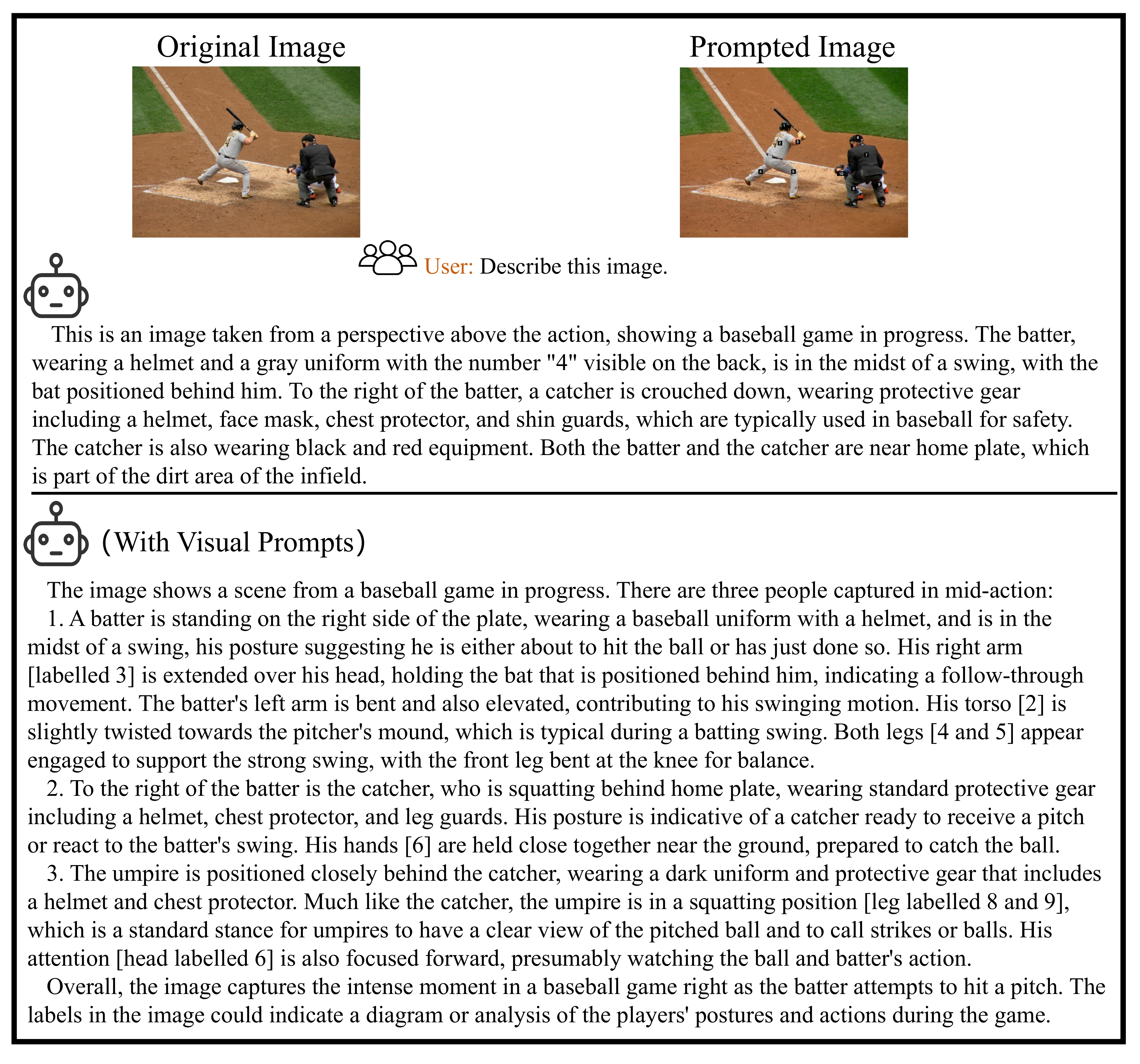}
\caption{
Supplying GPT4V with detailed visual prompts enhances its capability to comprehend and grasp fine-grained information in the image, thereby generating more detailed descriptive responses.
}
\label{fig:gpt4v_describtion}
\vspace{-2ex}
\end{figure*}

\begin{figure*}[tb]
\centering
\includegraphics[width=0.99 \linewidth]{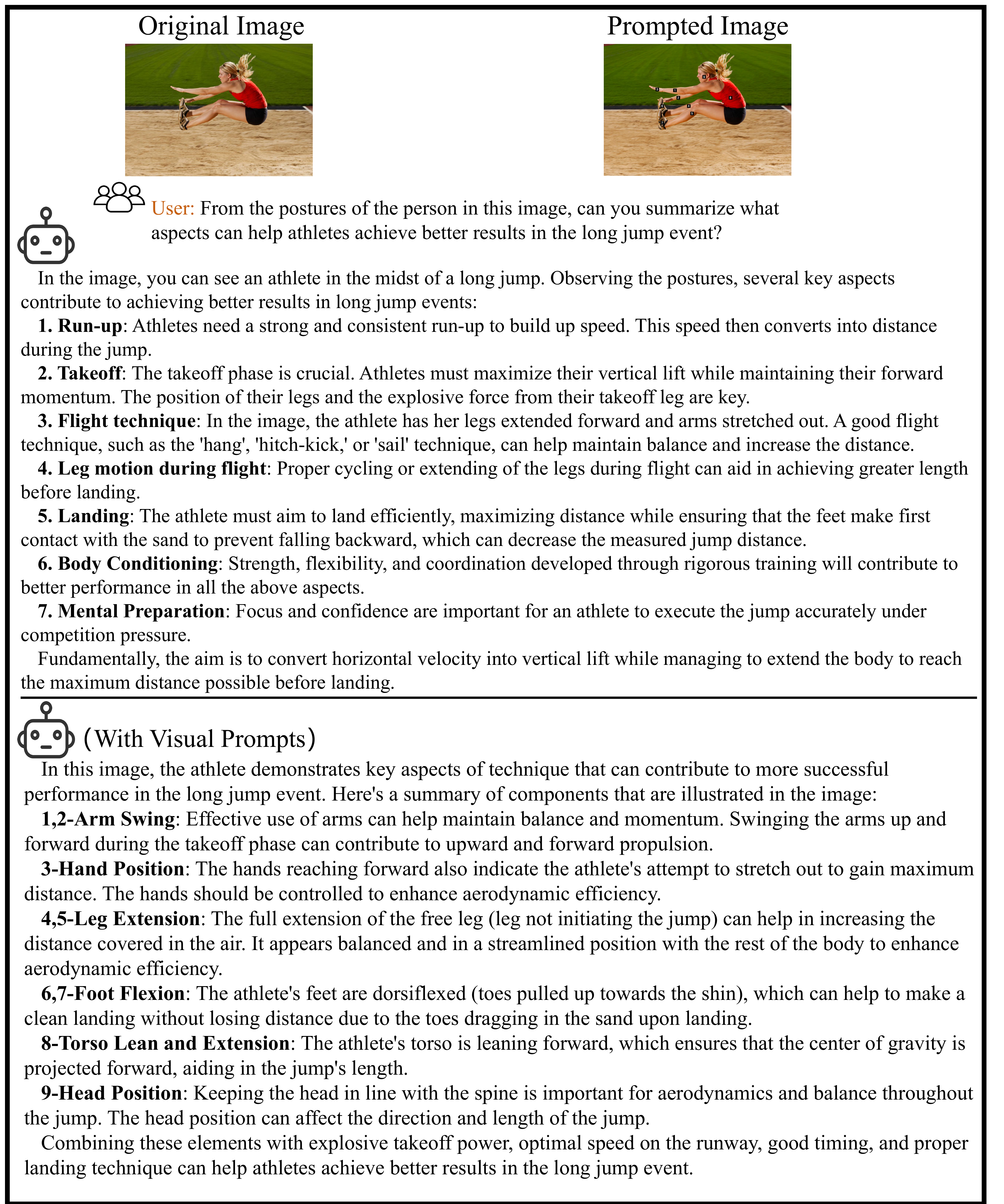}
\caption{
Providing fine-grained visual prompt labels to GPT4V enables it to generate more detailed responses. We observe that directly inputting unlabeled images, GPT4V may generate sentences unrelated to the image content, such as "Mental Preparation." However, upon providing visual prompt labels, GPT4V demonstrates an enhanced ability to comprehend additional details within the image, thus generating responses that are more in line with the image content.
}
\label{fig:gpt4v_action}
\vspace{-2ex}
\end{figure*}

Following SoM\cite{SoM}, we explore the effectiveness of fine-grained visual prompts in enhancing mLLMs'(such as GPT4V) ability to comprehend images, thus generating more satisfactory responses. 
We observe that, when provided with part-level information, mLLMs exhibit heightened attention to each part of the objects in the image, systematically describing them in sequence, various examples are shown in \cref{fig:gpt4v_ps5}, \cref{fig:gpt4v_face}, \cref{fig:gpt4v_describtion}, and \cref{fig:gpt4v_action}.

\setlength{\tabcolsep}{4pt}
\begin{table}[t]
\caption{\textbf{PACO part taxonomy} from \cite{ramanathan2023paco}.}
\begin{center}
\vspace{-2em}
\resizebox{0.98\columnwidth}{!}{
\begin{tabular}{r l l}
\toprule
Id & Object Categories &  Part Categories\\
\midrule
1 & ball & - \\
2  &  basket & bottom, handle, inner\_side, cover, side, rim, base\\
3  &  belt & buckle, end\_tip, strap, frame, bar, prong, loop, hole\\
4  &  bench & stretcher, seat, back, table\_top, leg, arm\\
5  &  bicycle & stem, fork, top\_tube, wheel, basket, seat\_stay, saddle, handlebar, pedal, gear, head\_tube, down\_tube, seat\_tube\\
6  &  blender & cable, handle, cover, spout, vapour\_cover, base, inner\_body, seal\_ring, cup, switch, food\_cup\\
7  &  book & page, cover\\
8  &  bottle & neck, label, shoulder, body, cap, bottom, inner\_body, closure, heel, top, handle, ring, sipper, capsule, spout, base, punt\\
9  &  bowl & inner\_body, bottom, body, rim, base\\
10  &  box & bottom, lid, inner\_side, side\\
11  &  broom & lower\_bristles, handle, brush\_cap, ring, shaft, brush\\
12  & bucket & handle, cover, body, base, inner\_body, bottom, loop, rim\\
13  & calculator & key, body\\
14  & can & pull\_tab, body, base, inner\_body, bottom, lid, text, rim\\
15  & car\_(automobile) & headlight, turnsignal, tank, windshield, mirror, sign, wiper, fender, trunk, windowpane, seat, logo, grille, antenna, hood, \\
 &  &  splashboard, bumper, rim, handle, runningboard, window, roof, wheel, taillight, steeringwheel\\
16  & carton & inner\_side, tapering\_top, cap, bottom, lid, text, side, top\\
17  & cellular\_telephone & button, screen, bezel, back\_cover\\
18  & chair & stretcher, swivel, apron, wheel, leg, base, spindle, seat, back, rail, stile, skirt, arm\\
19  & clock & cable, decoration, hand, pediment, finial, case, base\\
20  & crate & bottom, handle, inner\_side, lid, side\\
21  & cup & inner\_body, handle, rim, base\\
22 & dog & teeth, neck, foot, head, body, nose, leg, tail, ear, eye\\
23 & drill & handle, body\\
24 & drum\_(musical\_instrument) & head, rim, cover, body, loop, lug, base\\
25 & earphone & headband, cable, ear\_pads, housing, slider\\
26 & fan & rod, canopy, motor, blade, base, string, light, bracket, fan\_box, pedestal\_column\\
27 & glass\_(drink\_container) & inner\_body, bottom, body, rim, base\\
28 & guitar & key, headstock, bridge, body, fingerboard, back, string, side, pickguard, hole\\
29 & hammer & handle, face, head, grip\\
30 & handbag & zip, inner\_body, handle, bottom, body, rim, base\\
31 & hat & logo, pom\_pom, inner\_side, strap, visor, rim\\
32 & helmet & face\_shield, logo, inner\_side, strap, visor, rim\\
33 & jar & handle, body, base, inner\_body, bottom, lid, sticker, text, rim\\
34 & kettle & cable, handle, lid, body, spout, base\\
35 & knife & handle, blade\\
36 & ladder & rail, step, top\_cap, foot\\
37 & lamp & shade\_inner\_side, cable, pipe, shade, bulb, shade\_cap, base, switch, finial\\
38 & laptop\_computer & cable, camera, base\_panel, keyboard, logo, back, screen, touchpad\\
39 & microwave\_oven & inner\_side, door\_handle, time\_display, control\_panel, turntable, dial, side, top\\
40 & mirror & frame\\
41 & mouse\_(computer\_equipment) & logo, scroll\_wheel, body, right\_button, wire, side\_button, left\_button\\
42 & mug & handle, body, base, inner\_body, bottom, text, drawing, rim\\
43 & napkin & -\\ 
44 & newspaper & text\\
45 & pan\_(for\_cooking) & bottom, handle, inner\_side, lid, side, rim, base\\
46 & pen & cap, grip, barrel, clip, tip\\
47 & pencil & body, lead, eraser, ferrule\\
48 & pillow & embroidery\\
49 & pipe & nozzle, colied\_tube, nozzle\_stem\\
50 & plastic\_bag & inner\_body, handle, text, hem, body\\
51 & plate & top, bottom, inner\_wall, body, rim, base\\
52 & pliers & jaw, handle, joint, blade\\
53 & remote\_control & logo, back, button\\
54 & scarf & fringes, body\\
55 & scissors & handle, screw, finger\_hole, blade\\
56 & screwdriver & blade, handle, tip, shank\\
57 & shoe & toe\_box, tongue, vamp, outsole, insole, backstay, lining, quarter, heel, throat, eyelet, lace, welt\\
58 & slipper\_(footwear) & toe\_box, vamp, outsole, strap, insole, lining\\
59 & soap & neck, label, shoulder, body, sipper, capsule, spout, push\_pull\_cap, cap, base, bottom, closure, punt, top\\
60 & sponge & rough\_surface\\
61 & spoon & neck, handle, bowl, tip\\
62 & stool & seat, leg, step, footrest\\
63 & sweater & shoulder, sleeve, neckband, hem, body, yoke, cuff\\
64 & table & stretcher, drawer, inner\_wall, shelf, apron, wheel, leg, top, rim\\
65 & tape\_(sticky\_cloth\_or\_paper) & roll\\
66 & telephone & button, screen, bezel, back\_cover\\
67 & television\_set & bottom, button, side, top, base\\
68 & tissue\_paper & roll\\
69 & towel & body, terry\_bar, hem, border\\
70 & trash\_can & label, body, wheel, inner\_body, bottom, lid, pedal, rim, hole\\
71 & tray & bottom, inner\_side, outer\_side, rim, base\\
72 & vase & neck, handle, foot, body, mouth\\
73 & wallet & inner\_body, flap\\
74 & watch & buckle, case, dial, hand, strap, window, lug\\
75 & wrench & handle, head\\
\bottomrule
\end{tabular}}
\end{center}
\label{table:paco_categories}
\vspace{-2em}
\end{table}
\setlength{\tabcolsep}{1.4pt}

\end{document}